\documentclass{article}

\usepackage[preprint]{corl_2026} 

\usepackage{graphicx} 
\usepackage{subcaption}
\usepackage{siunitx}
\usepackage{amsmath} 
\usepackage{amssymb}  
\usepackage{algorithm}
\usepackage{algpseudocode}
\usepackage{booktabs}
\usepackage{tabularx}
\usepackage[normalem]{ulem}
\usepackage{acro}
\usepackage{placeins}
\newcommand{\citationneeded}{\textcolor{revisionblue}{[citation needed]}}

\newcommand{\blue}[1]{{#1}}
\newcommand{\green}[1]{{#1}}
\newcommand{\edit}[1]{#1}
\newcommand{\red}[1]{\textcolor{revisionred}{#1}}
\newcommand{\todo}[1]{\textcolor{revisionred}{#1}}
\newcommand{\white}[1]{\textcolor{revisionwhite}{#1}}
\newcommand{\figref}[1]{Figure \ref{#1}}
\newcommand{\algoref}[1]{Algorithm \ref{#1}}
\newcommand{\tableref}[1]{Table \ref{#1}}
\newcommand{\equationref}[1]{Equation~\ref{#1}}
\newcommand{\comment}[1]{\(\triangleright\) {#1}}
\newcommand{\delete}[1]{\sout{#1}}
\usepackage{multirow}
\usepackage{pgfplots}
\pgfplotsset{compat=1.16}
\definecolor{revisionblue}{rgb}{0,0,1}
\definecolor{revisionred}{rgb}{1,0,0}
\definecolor{revisiongreen}{rgb}{0,1,0}
\definecolor{revisionwhite}{rgb}{1,1,1}
\definecolor{black}{rgb}{0,0,0}
\definecolor{mydarkblue}{rgb}{0,0.08,0.45}

\DeclareAcronym{snn}{
  short = SNN,
  long  = soft nearest-neighbor,
}

\DeclareAcronym{bps}{
  short = BPS,
  long  = Basis Point Set,
}

\title{Continual Learning for 6-DoF Grasp Synthesis via Experience and Demonstrations}

\author{
  Giulio Schiavi\\
  Autonomous Systems Lab\\
  ETH Zurich, Switzerland\\
  \texttt{gschiavi@ethz.ch}
  \And
  Andrei Cramariuc\\
  Robotics Systems Lab \\
  ETH Zurich, Switzerland
  \And
  Michael Pantic\\
  Autonomous Systems Lab\\
  ETH Zurich, Switzerland
  \And
  Roland Siegwart\\
  Autonomous Systems Lab\\
  ETH Zurich, Switzerland
}

\begin{document}
\maketitle

\begin{figure}[h]
    \centering
    
    \includegraphics[width=0.9\linewidth]{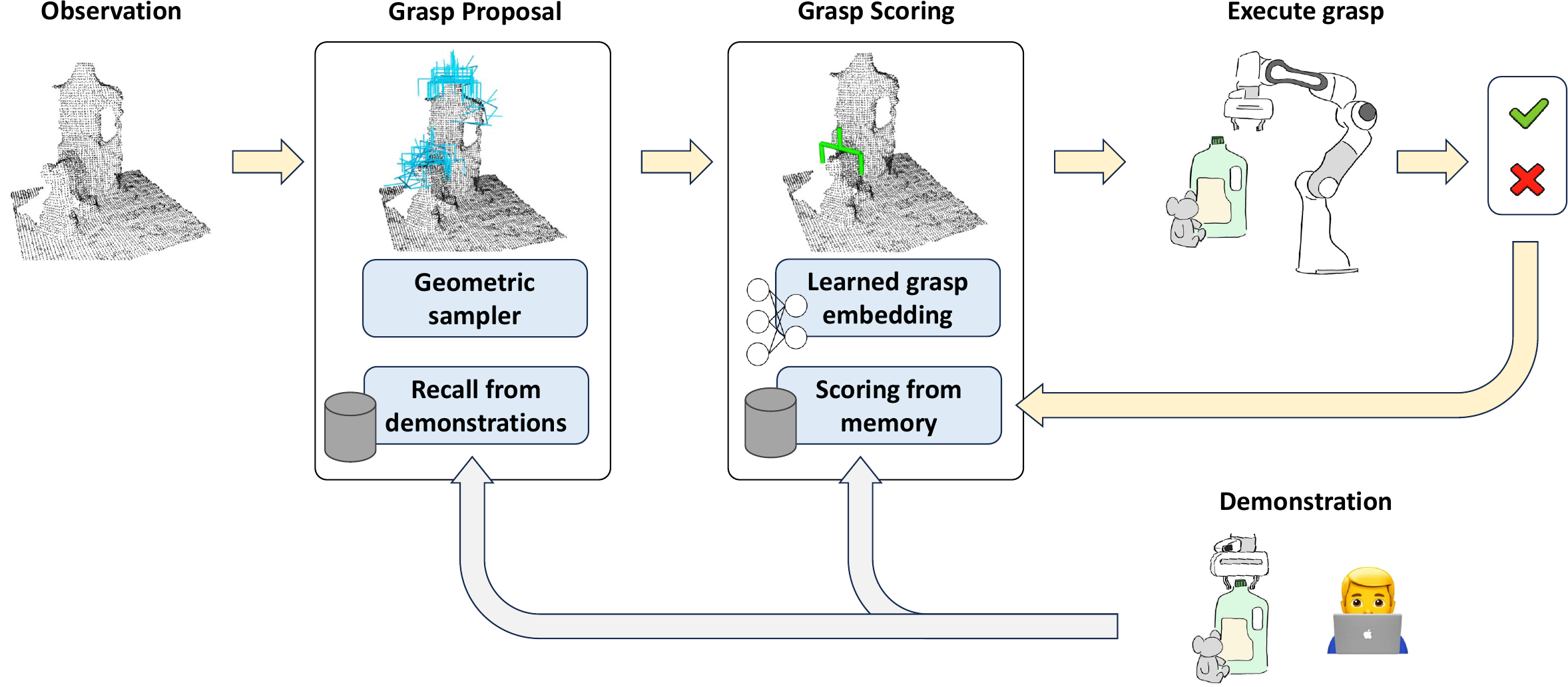}
    \caption{We present a continual-learning framework for 6-DoF grasp synthesis that is pretrained in simulation and keeps adapting during deployment. By leveraging memory in a learned embedding space, our method adapts from grasp outcomes and recalled user demonstrations without updating network weights.}
    \label{fig:banner}
    \vspace{-1.0em}
\end{figure}

\begin{abstract}
Most current grasp synthesis systems are trained offline and remain fixed during deployment. While this works well when deployment conditions resemble the training data, performance can degrade when robots encounter conditions they have not seen before, such as unfamiliar objects. In this work, we present a continual-learning framework for single-view 6-DoF grasp synthesis for a parallel-jaw gripper in cluttered scenes. Rather than finetuning a large parametric model, our method adapts through memory in a learned embedding space: grasp outcomes update future grasp scores, while optional user demonstrations are recalled and transferred to new scenes as additional candidate grasps. We evaluate our method in simulation and in extensive real-world experiments comprising over 1500 grasp trials. We show that our method matches the performance of existing 6-DoF grasping baselines even before adaptation, \green{improves online on unseen objects from categories absent or underrepresented during training}, and supports long-horizon continual learning with limited forgetting. In real-world experiments, our method reaches over 90\% success rates on several challenging object categories after only 50 online grasp attempts.
Videos and code at \url{https://giuschio.github.io/cl_grasping/}.
\end{abstract}

\keywords{Grasping, Continual Learning, Robot Manipulation}

\section{INTRODUCTION}
\label{sec:intro}
Current 6-DoF grasp synthesis systems can generate grasps in cluttered scenes with impressive reliability. Most do so by training parametric models on large offline datasets of synthetic or real-world grasp examples~\citep{fang2020graspnet,back_grasp_clutter6d,edgegraspnet,breyer2020volumetric,contact_graspnet}. Once trained, however, these models are typically frozen during deployment, and can still fail when the robot encounters conditions that are poorly covered by the training distribution, \green{such as unfamiliar object geometries that were absent or underrepresented during training}. One promising direction is to let grasping systems continue learning on site from the feedback available to them, such as grasp outcomes and occasional user demonstrations.

In this work, we present a continual-learning framework for 6-DoF grasp synthesis. As in many prior works~\citep{edgegraspnet, pas_gpd}, our system (\figref{fig:banner}) has two main components: a proposal module that generates candidate grasps and a scoring module that ranks them for execution. Unlike standard learned grasping systems, however, both modules can change during deployment through memory. Instead of using a parametric grasp scorer that remains fixed after training, we use a memory-based scorer in a learned embedding space, so new grasp outcomes immediately influence future scores. We also augment heuristic grasp proposals with lightweight recall from user demonstrations, transferring previously shown grasps to geometrically similar regions in new scenes. Together, these mechanisms let the system combine offline experience, deployment trials, and demonstrations into a single adaptive system. In addition, because adaptation only requires adding entries to memory, each update is simple and practical to apply on site, without requiring gradient-based retraining. 

We evaluate our method in simulation and on a real robot and compare it with common 6-DoF grasping baselines. Our method achieves strong offline performance, \green{improves online through deployment experience on unfamiliar objects}, and supports long-horizon continual learning with limited forgetting. In real-world experiments comprising over 1500 grasp trials, online adaptation improves performance across challenging object categories and reaches over 90\% success on several of them.

\section{RELATED WORK}
\label{sec:related_work}

\subsection{Deep Grasp Synthesis}
6-DoF grasp synthesis aims to predict feasible SE(3) grasp poses from scene observations. A common strategy is a proposal-and-score pipeline: generate a set of candidate grasps, then rank them according to predicted grasp success. Some methods sample candidates with geometric heuristics, such as GPD~\citep{pas_gpd} and EdgeGraspNet~\citep{edgegraspnet}. Others evaluate dense anchors over points or voxels, as in VGN~\citep{breyer2020volumetric}, ICGNet~\citep{icg_net}, ContactGraspNet~\citep{contact_graspnet}, GraspNet~\citep{fang2020graspnet}, and AnyGrasp~\citep{fang2023anygrasp}. Recent progress has come from larger datasets and stronger priors, such as adversarial sim-to-real alignment~\citep{gpdan_robust_grasping}, antipodal constraints~\citep{ma_generalization_prior_knowledge}, and center-of-gravity priors~\citep{fang2023anygrasp}. Other methods learn the proposal distribution with VAEs~\citep{fox_6dofgraspnet} or diffusion models~\citep{murali2025graspgen, song_implicit_grasp_diffusion, barad2023graspldm}. Despite this progress, current systems remain tied to the coverage of their training data, and can still fail when deployment scenes differ from the examples seen offline. 

\subsection{Adaptation and Continual Learning for Grasping}
Prior work has addressed the problem of how grasping models should be adapted during deployment. Existing approaches use self-training signals~\citep{gilles_continual_vacuum, liu_continual_grasping}, continual-learning regularizers~\citep{ren_digital_twin}, reinforcement learning~\citep{shi_uncertainty_driven_exploration, le_curriculum_learning}, or retraining with deployment data~\citep{julian_never_stop_learning}. Other works exploit structural invariances in 2D top-down grasping to adapt from few examples~\citep{yang_attribute_based}. These methods show that deployment data is valuable, but often consider settings with simplifying assumptions, such as top-down grasping or access to test-time CAD models. \blue{We instead consider single-view 6-DoF grasping in scenes with multiple objects. Our approach also differs in how adaptation is performed: rather than updating the parameters of the grasping model, we incorporate deployment experience non-parametrically by adding entries to memory. In our experiments, we compare our approach against parametric finetuning using the protocol of~\citet{julian_never_stop_learning}.}

\subsection{Registration- and Retrieval-Based Grasping}
Retrieval-based methods store grasps and transfer them to new scenes through registration or correspondences. Early work transfers grasps geometrically~\citep{kopicki_registration}. Later methods integrate category templates~\citep{wen2022transgrasp}, semantic correspondences~\citep{ju2025robo}, or continual retrieval from successful grasps~\citep{patten_dgcmnet}. These frameworks can often incorporate new demonstrations without retraining, but they depend on database coverage and can be sensitive to registration errors. \blue{Our method differs in three main ways. First, recall adds to geometric sampling instead of replacing it; the system can still propose grasps for unfamiliar objects. Second, whereas retrieval methods learn from successful grasps or demonstrations only, our method can learn from both successful and failed grasp attempts. Third, registration is used only to generate candidates, not to determine their scores. This allows us to reject a well-registered recall when the accumulated outcome evidence indicates that the transferred grasp is unlikely to succeed. In our experiments, we isolate these differences by comparing against a recall-only pipeline that disables geometric sampling and adaptive scoring.}

\subsection{Exploratory and Trial-Based Grasp Learning}
Exploratory and trial-based grasp learning methods are closely related to our work. Dex-Net~\citep{dexnet10}, BORGES~\citep{danielczuk_exploratory_grasping}, and LEGS~\citep{legs_exploratory_grasping} maintain probabilistic grasp success estimates and update them from heuristic evaluations or real-world trials. These works primarily target efficient grasp dataset generation~\citep{dexnet10} or exploratory grasping for a single isolated object~\citep{legs_exploratory_grasping}. They also tend to treat grasps independently, maintaining separate success estimates for each candidate grasp or object pose. We build on the same trial-based idea, but use it for 6-DoF grasping from partial observations in cluttered scenes. Rather than maintaining independent beliefs over individual grasps, we operate in a learned embedding space, so evidence from one grasp attempt can generalize to similar grasps.

\section{METHOD}
\label{sec:method}

We use a 7-DOF robotic arm with a depth camera and a parallel-jaw gripper operating in a 30 × 30 \si{cm} tabletop workspace. From a single depth observation, captured from a randomly sampled viewpoint, the robot must generate feasible 6-DoF grasp candidates. Our pipeline is organized into two main modules: a proposal module which generates grasp candidates from the current observation, and a scoring module which ranks them for execution according to their estimated success likelihood. The full pipeline is shown in~\figref{fig:banner}.

\subsection{Grasp Proposal}
\label{sec:grasp_sampling}
\label{sec:proposal_recall}
At each inference step, we construct the proposal set (i.e. the set of candidate grasps that will be scored later) from two sources: geometric sampling and demonstration recall. This lets us generate a broad set of grasp candidates for any scene, while leveraging demonstrations to add object-specific grasp modes that might be missed by the fixed heuristics. Our geometric sampler follows the contact-based 6-DoF proposal strategy which is common in prior work~\citep{edgegraspnet, contact_graspnet, icg_net}: it samples approach and contact points from the current pointcloud and uses the local surface normal to define the gripper orientation. Alongside these contact-based grasps, we also include top-down proposals with a fixed vertical approach direction, which we empirically find useful on smaller objects for which depth observations give unreliable side normals. For demonstration recall, we query a recall memory in which each entry consists of a demonstrated grasp and the local spherical pointcloud patch observed around it. We match these stored patches to regions in the current scene using FPFH features~\citep{fpfh_features} followed by point-to-plane ICP registration~\citep{icp_cit}. When registration succeeds, the estimated transform maps the demonstrated grasp into the current scene, adding it as a new proposal. Finally, we filter all sampled and recalled proposals for kinematic feasibility and collision.

\subsection{Grasp Scoring}
\label{sec:memory_scoring}
\label{sec:grasp_embedding}

After sampling, we score every candidate (both sampled and recalled). We do this in two steps. First, an encoder maps each grasp to a low-dimensional embedding space. 
Then, a memory-based scorer estimates the success probability of each grasp by aggregating nearby labeled data points, including data from offline training, real deployment grasp outcomes and demonstrations. The advantage of this approach over an end-to-end parametric model is that it can be updated instantly by adding new data to the memory.
\subsubsection{Encoder Training}
We represent each grasp by a local pointcloud patch expressed in the grasp frame.
We serialize the patch with a \ac{bps}~\citep{prokudin2019efficientlearningpointclouds} and pass it through an MLP encoder.
Let $x_i$ denote the serialized pointcloud for grasp $i$, let $y_i \in \{0,1\}$ denote its binary success label, and let $z_i = E_\phi(x_i)$ denote its embedding, where $E_\phi$ is the encoder. The encoder outputs a 32-dimensional embedding, which we normalize to lie on the unit sphere. 

Following prior work on training embeddings for non-parametric classification~\citep{hinton_snn_2007}, we train the encoder on a mixed \ac{snn} and reconstruction loss.
The \ac{snn} loss~\citep{hinton_snn_2007}, computed over a batch of size $B$, encourages grasps with the same outcome to cluster together:
\begin{equation}
\mathcal{L}_{\mathrm{SNN}}
= - \frac{1}{B} \sum_{i=1}^B
\log
\frac{\sum_{j\neq i} \exp\!\left(-\frac{\|z_i - z_j\|^2}{\tau_{\mathrm{train}}}\right) \mathbf{1}[y_j = y_i]}
     {\sum_{j\neq i} \exp\!\left(-\frac{\|z_i - z_j\|^2}{\tau_{\mathrm{train}}}\right)}.
\label{eq:snn_loss}
\end{equation}
Here $\tau_{\mathrm{train}}$ is a temperature parameter. The reconstruction loss, parametrized through an auxiliary decoder $D_\psi$, keeps the embedding tied to local geometry:
\begin{equation}
\mathcal{L}_{\mathrm{rec}}
= \frac{1}{B} \sum_{i=1}^B \| x_i - D_\psi(z_i) \|_2.
\label{eq:reconstruction_loss}
\end{equation}
We first pretrain the encoder with $\mathcal{L}_{\mathrm{rec}}$, and then optimize the mixed loss $\mathcal{L}_{\mathrm{embed}} = \alpha\,\mathcal{L}_{\mathrm{SNN}} + (1-\alpha)\,\mathcal{L}_{\mathrm{rec}}$ (we use $\alpha=0.8$). Additional embedding and scorer implementation details are provided in \blue{Appendix~\ref{appendix:embedding_details}}, while \blue{Appendix~\ref{appendix:embedding_comparison}} compares this \ac{snn}-trained embedding against one trained using a binary cross-entropy loss.

\subsubsection{Memory-based scorer}
To score each grasp, we aggregate local evidence in the learned embedding space produced by the encoder.
We maintain a scoring memory $\mathcal{M} = \{(z_i,y_i,\rho_i)\}_{i=1}^{N}$, where each entry contains a grasp embedding $z_i$, a binary outcome label $y_i \in \{0,1\}$, and a base evidence weight $\rho_i$. We initialize the scoring memory with offline training data and then expand it during deployment with grasp outcomes and demonstrations.
At inference, given a query grasp $g_q$, we compute its embedding $z_q$ and retrieve its $K$ nearest labeled neighbors from the scoring memory. We model our belief over the success probability $s_q$ with a Beta posterior $p(s_q \mid \mathcal{M}) = \mathrm{Beta}(s_q;\alpha_q,\beta_q)$, where each neighbor contributes evidence in the form of pseudo-counts. Each neighbor contributes a pseudo-count weighted by its base weight and by an RBF kernel of its distance to the query,
\begin{equation}
w_j = \rho_j \exp\!\left( -\frac{\|z_q - z_j\|^2}{\tau_{\mathrm{eval}}} \right),
\label{eq:rbf_kernel}
\end{equation}
where $\tau_{\mathrm{eval}}$ is a bandwidth parameter. For each query, we cap the total contribution of offline neighbors at $W_{\max}^{\mathrm{off}}$ pseudo-counts. This prevents the offline part of the scoring memory from dominating the posterior and slowing adaptation, while preserving the local offline success-to-failure ratio. 
This gives the posterior parameters
\begin{equation}
\begin{aligned}
\alpha_q = \alpha_0
+ \kappa^{\mathrm{off}}
\sum_{j \in \mathcal{N}_{\mathrm{off}}}
w_j y_j
+ \sum_{j \in \mathcal{N}_{\mathrm{on}}}
w_j y_j,
\end{aligned}
\label{eq:beta_evidence_alpha}
\end{equation}
\begin{equation}
\begin{aligned}
\beta_q = \beta_0
+ \kappa^{\mathrm{off}}
\sum_{j \in \mathcal{N}_{\mathrm{off}}}
w_j (1-y_j)
+ \sum_{j \in \mathcal{N}_{\mathrm{on}}}
w_j (1-y_j).
\end{aligned}
\label{eq:beta_evidence_beta}
\end{equation}
where $\mathcal{N}_{\mathrm{off}}$ and $\mathcal{N}_{\mathrm{on}}$ denote the offline and online neighbors of $g_q$, $\kappa^{\mathrm{off}}$ is the offline rescaling factor, and $\alpha_0$ and $\beta_0$ are priors that define the default score for grasps with little nearby evidence. The grasp score is the posterior mean,
\begin{equation}
\mathbb{E}[s_q] = \frac{\alpha_q}{\alpha_q+\beta_q}
\label{eq:posterior_mean}
\end{equation}
which increases when nearby evidence is mostly successful and decreases when it is mostly failed. \blue{We use different base weights $\rho_i$ for offline and online evidence, with higher weight on online outcomes to improve the speed of adaptation. Demonstrations are weighted adaptively from the current posterior odds, choosing enough positive evidence for the demonstrated grasp to exceed the current best grasp by a margin, while enforcing a minimum weight so every demonstration affects the scorer. Appendix~\ref{appendix:evidence_weights} further discusses how these weights are tuned.}

\subsection{Online Operation}
\label{sec:online_operation}
We use simulation data, collected using geometric heuristic grasp sampling, to train the encoder and initialize the memory-based scorer, which gives the robot a strong prior before deployment-time adaptation. During deployment, we add each binary grasp outcome to the scoring memory, so trial-and-error immediately informs the scoring of future candidates. \blue{Labels are automatic: on the real robot, we read the gripper finger positions and label a grasp as successful if the resulting finger width is greater than zero.} \blue{User demonstrations provide an optional second source of adaptation.} When a demonstration is available, we add it to the scoring memory with an adaptive weight chosen from the current posterior, so that the demonstrated grasp exceeds the current best candidate by a fixed margin. This encodes the intuition that user-demonstrated grasps should be preferred in sufficiently similar conditions, while still letting the required evidence depend on the current local memory. We also add the demonstration to the recall memory, allowing it to be transferred to future scenes with similar local geometry. \blue{The system can therefore adapt autonomously from grasp outcomes alone, while demonstrations can additionally introduce grasp modes missing from the geometric sampler.}

\section{Experiments}
\label{sec:experiments}
We evaluate our method in simulation and on a real robot. The experiments first test the base model trained from simulation data, then measure how deployment experience changes performance on \green{unseen objects from absent or underrepresented categories}, both with category-specific adaptation and in a continual-learning setting. 

\subsection{Simulation Experiments}
\label{sec:sim_experiments}
We use the PyBullet simulation benchmark introduced by VGN and later adopted by EdgeGraspNet and ICGNet~\citep{breyer2020volumetric, edgegraspnet, icg_net}. Following the packed setting from VGN, each cluttered scene contains one to five upright objects placed close together on a table. We train on the VGN object set, which is also used by EdgeGraspNet and ICGNet.
For evaluation, we replace the training objects with \green{443 unseen objects} from DexGraspNet~\citep{wang2022dexgraspnet}, grouped into ten semantic categories: Airplane, Animal, Bottle, Bowl, Drill, Fasteners, Hammer, Mug, Pliers, and Screwdriver. \blue{We chose categories underrepresented in training but representative of objects a robot might encounter. All instances are unseen. We give a more detailed analysis of the test object distribution in Appendix~\ref{appendix:evaluation_novelty}.}

\paragraph{Baseline Performance and Single-Category Adaptation}
\label{sec:sim_category_adaptation}
We first compare the offline version of our model against existing 6-DoF grasping baselines (namely ICGNet~\citep{icg_net}, ContactGraspNet~\citep{contact_graspnet}, 
EdgeGraspNet~\citep{edgegraspnet}), before any deployment-time adaptation. As shown in~\tableref{tab:category_level_performance_frozen}, our method matches the strongest baseline, EdgeGraspNet, across the tested categories and slightly improves the average success rate. The non-parametric scorer therefore does not trade away offline performance.

We then adapt to each object category independently. For each category, the robot collects 100 grasp attempts and receives at most 5 user demonstrations. \blue{First, we remove demonstration recall and adapt the scorer using automatically observed grasp outcomes only. This isolates outcome-based scoring adaptation and measures how much the system can improve autonomously through trial and error.} Second, we remove geometric sampling and the memory-based scorer, so grasps are generated only by recalling demonstrations. \blue{This recall-only ablation serves as a proxy for methods that transfer grasps to new scenes through geometric retrieval and registration~\citep{patten_dgcmnet,kopicki_registration,wen2022transgrasp}.} Third, we finetune EdgeGraspNet using the same 100 per-category grasp labels and a balanced loss that weights category-specific data and the original EdgeGraspNet training data equally. \blue{This provides a parametric adaptation baseline under the same 6-DoF grasping setting and deployment-data budget. The 50/50 mixture follows the finetuning protocol of~\citet{julian_never_stop_learning}.}

As shown in~\tableref{tab:category_level_performance_adaptation}, online adaptation improves performance on \green{all evaluated categories}. Our full method reaches 98.1\% average success, up from 94.6\% for the base model, and achieves at least 98\% success on 7 of 10 categories. Adapting the scorer alone already performs well, reaching 97.1\% success rate on average. Recall-only performs worse because each category contains many objects and configurations, while only a small number of demonstrations is available. This supports our use of recall as a supplement to geometric sampling rather than as the sole proposal mechanism. Finetuning yields only a small gain, from 92.9\% to 93.5\%, suggesting that 100 category-specific labels are not enough to effectively finetune a large parametric model, even with access to the original training data. The contrast is also computational: our method adapts by adding memory entries at negligible cost, while finetuning takes one hour per category on an NVIDIA RTX 4070 GPU.

\begin{table*}[t]
\centering
\resizebox{\textwidth}{!}{
\begin{tabular}{lllllllllllll}
\toprule
Method & Train Obj. & Airplane & Animal & Bottle & Bowl & Drill & Fasteners & Hammer & Mug & Pliers & Screwdriver & \multicolumn{1}{c}{\shortstack{Average\\(excl. training)}} \\
\midrule
ICGNet          & 91.6\% & 75.4\% & 89.0\% & 95.4\% & 52.1\% & 95.8\% & 81.8\% & 88.1\% & 85.2\% & 56.3\% & 67.0\% & 78.6\% \\
ContactGraspNet \white{---------} & 89.2\% & 83.0\% & 89.5\% & \textbf{96.6\%} & \textbf{95.4\%} & 93.1\%& 90.6\% & 74.1\% & 81.6\% & 39.7\% & 74.5\% & 81.8\% \\

EdgeGraspNet    & 95.3\% & 91.6\% & 89.0\% & 96.3\% & 91.0\% & \textbf{96.8\%} & 91.8\% & \textbf{96.0\%} & 91.2\% & 93.1\% & 92.6\% & 92.9\% \\
Ours (base)     & \textbf{98.1\%} & \textbf{92.8\%} & \textbf{93.8\%} & 96.4\% & 88.6\% & 93.5\% & \textbf{97.4\%} & \textbf{96.0\%} & \textbf{92.4\%} & \textbf{98.0\%} & \textbf{97.2\%}& \textbf{94.6\%} \\
\bottomrule
\end{tabular}
}
\caption{\textbf{Grasp success rates without online adaptation (simulation).} For each object category and method, results are computed over 2,500 grasp attempts. Our base model performs better than or comparably to the baselines across all categories, while achieving the highest average performance.}
\label{tab:category_level_performance_frozen}

\vspace{0.5em}

\resizebox{\textwidth}{!}{
\begin{tabular}{lllllllllllll}
\toprule
Method & Train Obj. & Airplane & Animal & Bottle & Bowl & Drill & Fasteners & Hammer & Mug & Pliers & Screwdriver & \multicolumn{1}{c}{\shortstack{Average\\(excl. training)}} \\
\midrule

EdgeGraspNet (finetuned) & - & 91.5\% & 90.3\% & 96.2\% & 92.4\% & 96.8\% & 92.0\% & 96.1\% & 94.7\% & 91.9\% & 92.7\% & 93.5\% \\
Ours (recall only)  & - & 87.4\% & 80.8\% & 81.2\% & 93.8\% & 90.5\% & 94.8\% & 85.2\% & 90.5\% & 59.3\% & 92.1\% & 85.6\% \\
Ours (scoring only)  & - & 96.0\% & 93.3\% & 98.2\% & \textbf{98.2\%} & 97.8\% & 97.7\% & \textbf{98.6\%} & 95.6\% & 97.6\% & 98.0\% & 97.1\% \\
Ours (full)  & - & \textbf{97.8\%} & \textbf{94.5\%} & \textbf{99.7\%} & 97.6\% & \textbf{98.1\%} & \textbf{98.0\%} & 98.5\% & \textbf{98.5\%} & \textbf{99.2\%} & \textbf{98.6\%} & \textbf{98.1\%} \\
\bottomrule
\end{tabular}
}
\caption{\textbf{Category-specific adaptation (simulation).} We adapt to each category independently and evaluate each method over 2,500 grasp attempts per category. Our method improves performance on all categories, raising the average from 94.6\% to 98.1\% and reaching 98\% on 7 of 10 categories. The scorer-only ablation performs well, while recall-only adaptation is less reliable. Finetuning EdgeGraspNet yields only a small gain.}
\label{tab:category_level_performance_adaptation}

\end{table*}

\paragraph{Continual Learning}
\label{sec:sim_continual_learning}
\begin{table}[t]
\centering
\resizebox{\textwidth}{!}{
\begin{tabular}{lllllllllllll}
\toprule
Method & Train Obj. & Airplane & Animal & Bottle & Bowl & Drill & Fasteners & Hammer & Mug & Pliers & Screwdriver & \multicolumn{1}{c}{\shortstack{Average\\(excl. training)}} \\
\midrule

Ours (base)     & 98.1\% & 92.8\% & 93.8\% & 96.4\% & 88.6\% & 93.5\% & 97.4\% & 96.0\% & 92.4\% & 98.0\% & 97.2\%& 94.6\% \\

Ours (+ Airplane)
& 97.8\%  & 97.0\% & - & - & - & - & - & - & - & - & - & - \\

Ours (+ Animal)
& 98.1\% & 96.2\%  & 95.7\% & - & - & - & - & - & - & - & - & - \\

Ours (+ Bottle)
& 98.9\%& 96.2\% & 95.0\% & 99.5\%  & - & - & - & - & - & - & - & - \\

Ours (+ Bowl)
& 98.7\% & 96.8\% & 96.0\%  & 99.5\% & 98.9\% & - & - & - & - & - & - & - \\

Ours (+ Drill)
& 98.7\% & 96.0\% & 95.0\% & 97.5\% & 98.3\%  & 97.9\% & - & - & - & - & - & - \\

Ours (+ Fasteners)
& 99.1\% & 96.5\% & 95.9\% & 98.7\% & 98.7\% & 98.6\% & 99.1\% & - & - & - & - & - \\

Ours (+ Hammer)
& 98.7\% & 95.2\% & 94.6\% & 98.5\% & 98.2\% & 98.1\% & 98.5\% & 98.0\% & - & - & - & - \\

Ours (+ Mug)
& 98.2\% & 96.0\%  & 93.3\%  & 98.6\% & 96.9\% & 98.0\% & 98.7\% & 98.5\% & 99.2\% & - & - & - \\

Ours (+ Pliers)
& 98.7\% & 96.2\% & 94.8\%  & 98.8\% & 97.5\%  & 98.0\%& 98.8\%  & 99.0\%  & 99.0\% & 99.2\% & - & - \\

Ours (+ Screwdriver)
& 98.1\%  & 95.0\% & 94.5\% & 99.2\%  & 98.0\%& 98.2\%  & 98.4\%  & 98.6\% & 98.3\%  & 98.4\% & 99.1\%  & 97.8\% \\

\bottomrule
\end{tabular}
}
\vspace{0.5em}
\caption{\textbf{Sequential continual learning in simulation.}
\green{The robot adapts to the ten evaluation categories one after another.} After each stage, we evaluate on the current category and all previously seen categories, using 1000 grasp attempts per category. \green{Our method adapts across categories without forgetting previous ones.} Final performance on average remains close to category-specific adaptations (97.8\% vs. 98.1\%, see~\tableref{tab:category_level_performance_adaptation}). 
}
\label{tab:continual_ours}
\vspace{-2.0em}
\end{table}

We next test the same adaptation mechanism in a longer continual-learning sequence. The robot adapts to all ten evaluation categories in the order shown in Table~\ref{tab:continual_ours} (the order is chosen arbitrarily). Each stage follows the same protocol as the previous experiment: 100 adaptation attempts on the current category, with a user demonstration after every failed attempt and a maximum of 5 demonstrations. Table~\ref{tab:continual_ours} reports the results. Our approach successfully adapts to each category sequentially, while previous categories remain strong rather than being forgotten. At the end of the sequence, our model has adapted to \green{ten evaluation categories covering 443 unseen objects}. The worst performance drop due to forgetting is 2.4 percentage points, and the final model outperforms the base model on every object category. Memory and runtime costs remain small: after adaptation, the memory footprint is 1.35 MB and the full planning cycle remains below 1.5 s. \blue{Appendix~\ref{appendix:runtime_memory} analyzes how the runtime and memory costs of our method scale with accumulated experience.}

\subsection{Real-World Experiments}
\label{sec:real_experiments}
\begin{figure*}[t]
    \centering
    \begin{minipage}[t]{0.135\textwidth}
        \vspace{0pt}
        \centering
        \includegraphics[width=\linewidth]{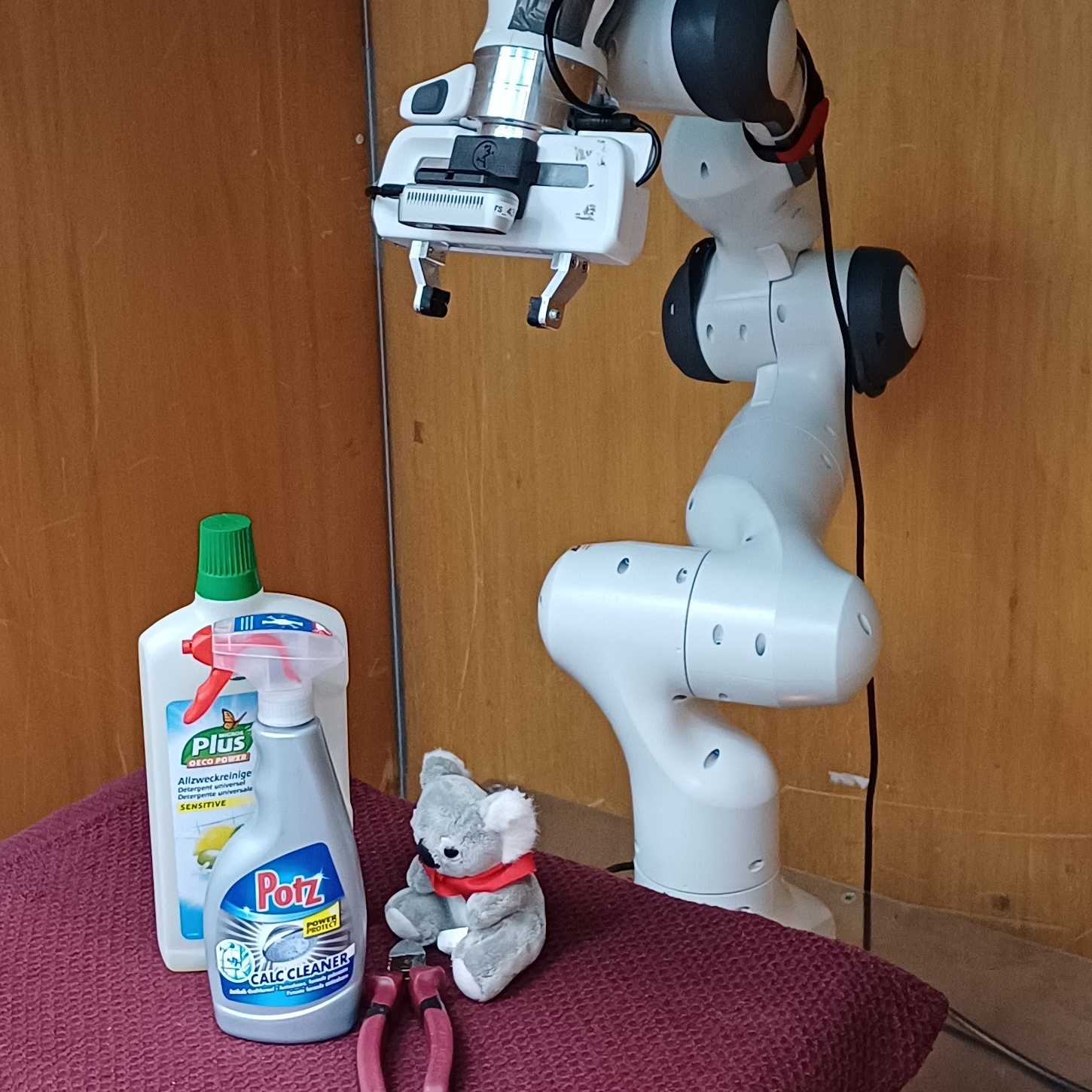}
        \par{\scriptsize Robot setup}
    \end{minipage}\hfill
    \begin{minipage}[t]{0.135\textwidth}
        \vspace{0pt}
        \centering
        \includegraphics[width=\linewidth]{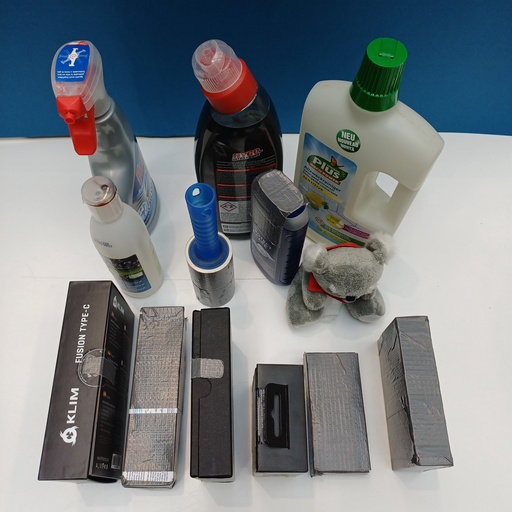}
        \par{\scriptsize Control}
    \end{minipage}\hfill
    \begin{minipage}[t]{0.135\textwidth}
        \vspace{0pt}
        \centering
        \includegraphics[width=\linewidth]{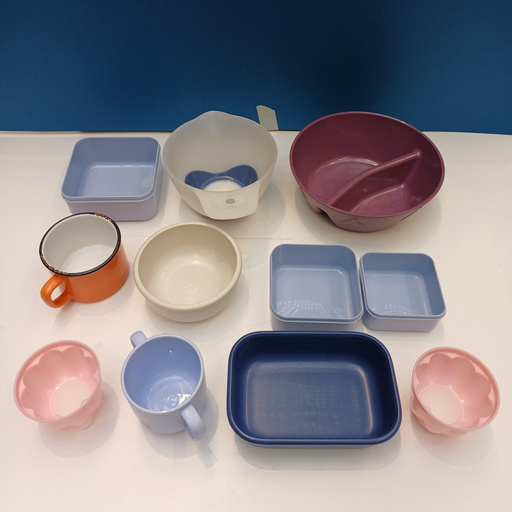}
        \par{\scriptsize Mugs/bowls}
    \end{minipage}\hfill
    \begin{minipage}[t]{0.135\textwidth}
        \vspace{0pt}
        \centering
        \includegraphics[width=\linewidth]{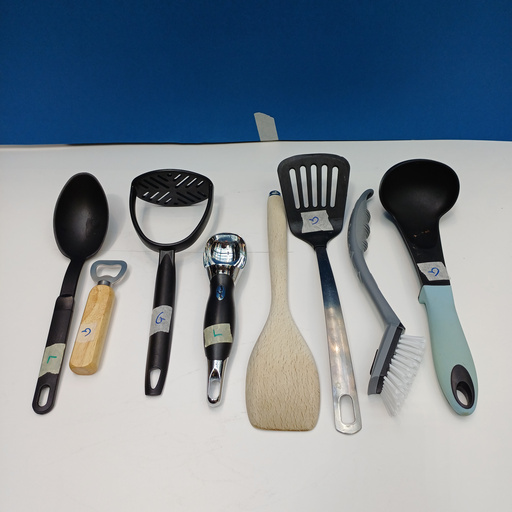}
        \par{\scriptsize Kitchen}
    \end{minipage}\hfill
    \begin{minipage}[t]{0.135\textwidth}
        \vspace{0pt}
        \centering
        \includegraphics[width=\linewidth]{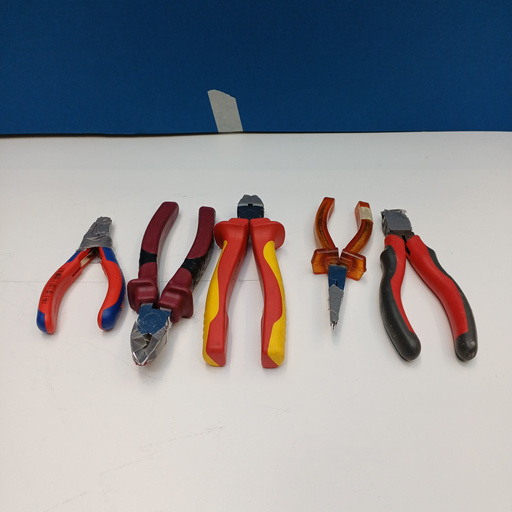}
        \par{\scriptsize Pliers}
    \end{minipage}\hfill
    \begin{minipage}[t]{0.135\textwidth}
        \vspace{0pt}
        \centering
        \includegraphics[width=\linewidth]{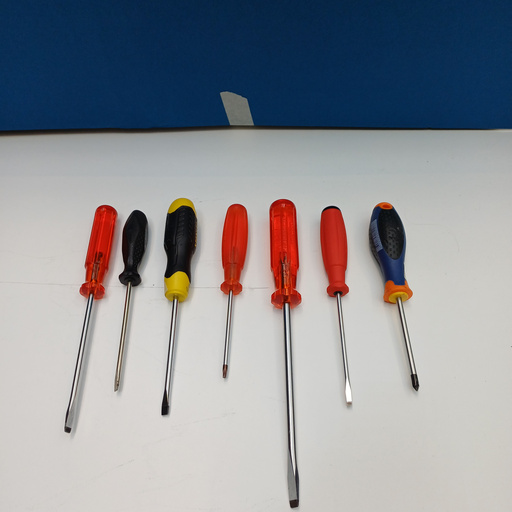}
        \par{\scriptsize Screwdrivers}
    \end{minipage}\hfill
    \begin{minipage}[t]{0.135\textwidth}
        \vspace{0pt}
        \centering
        \includegraphics[width=\linewidth]{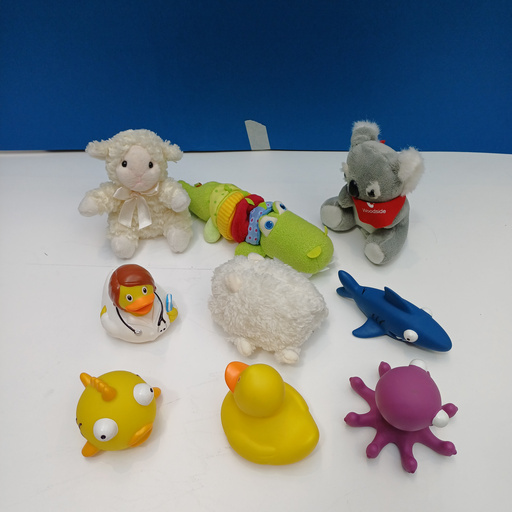}
        \par{\scriptsize Toys}
    \end{minipage}
    \caption{
    \textbf{Real-world setup and object sets used for evaluation.}
    The control objects are chosen to resemble common grasp-synthesis evaluation objects, while the remaining categories include mugs and bowls, kitchen tools, pliers, screwdrivers, and toys with more challenging geometries, materials, thin structures, and non-uniform mass distributions. \blue{The single-camera scenes have moderate occlusion. Stacked or interlocked clutter is outside our scope.}
    }
    \label{fig:real_objects}
    
\end{figure*}

Next, we move to the real robot. Our setup uses a Franka Panda arm~\citep{frankapandacit}, a parallel-jaw gripper, and a RealSense D435i depth camera. We use 54 objects: one control set of 13 objects and five challenge sets, with 11 mugs and bowls, 9 kitchen tools, 5 pliers, 7 screwdrivers, and 9 toys (\figref{fig:real_objects}). The control objects are modeled on the evaluation set used by ICGNet~\citep{icg_net}, and roughly match the training distribution. The challenge sets stress real-world effects such as thin geometry, material variation, and uneven mass distributions. We construct cluttered scenes using the same protocol as in simulation. \blue{We evaluate non-targeted clearing without semantic segmentation.} At each step, the robot executes the highest-scoring collision-free grasp, removes the object if successful, and repeats until the scene is cleared or terminated. We report grasp success rate (i.e. the fraction of attempts that succeed) and scene-clear rate (i.e. the fraction of scenes fully cleared). If no feasible grasp is found, the robot replans from the next viewpoint in a fixed fallback sequence, up to three times, reducing sensitivity to unfavorable initial views while preserving deterministic comparisons across methods. If all three viewpoints fail, we terminate the scene and count one failed attempt. In total, our evaluation comprises more than 1500 real-world grasp attempts.

\paragraph{Baseline Performance and Single-Category Adaptation}
We first adapt to each real-world category independently. For each category, we allow up to 50 grasp attempts for adaptation and provide a user demonstration after each failed attempt. We then evaluate our model before and after adaptation, and compare it against EdgeGraspNet, AnyGrasp~\citep{fang2023anygrasp}, and M2T2~\citep{yuan2023m2t2}. For each category, we construct 20 evaluation scenes and use the same set of scenes for all methods. Because each scene contains multiple objects, each category-level evaluation corresponds to several dozen executed grasps per method\footnote{The exact number of attempts differs across methods and evaluations because the number of grasps needed to clear each scene depends on the method and difficulty of the object set.}. The first six columns of~\tableref{tab:real_category_adaptation} summarize the category-level results. \blue{On the control set, our base model is comparable to the strongest baselines, reaching 80.0\% success compared with 80.2\% for EdgeGraspNet and 78.8\% for AnyGrasp. Across the challenge sets, AnyGrasp is generally the strongest external baseline, but our base model outperforms all baselines on four of the five sets, with AnyGrasp slightly ahead on toys. The gap is largest on pliers and screwdrivers, where AnyGrasp, the strongest baseline on both, reaches 28.6\% and 54.9\% success, compared with 62.5\% and 85.7\% for our base model. EdgeGraspNet and M2T2 perform substantially worse on these two sets.} EdgeGraspNet exhibits a particularly clear proposal-generation failure on pliers and screwdrivers, where it achieves 0.0\% success and 0.0\% scene-clear rate. On these thin objects, it rarely proposes any collision-free grasps, and the few valid proposals fail. We attribute this failure mode to its contact-normal-based grasp-candidate sampling, a common strategy in current methods~\citep{icg_net,edgegraspnet,contact_graspnet}. This heuristic works well for bulky objects with clear side surfaces, but struggles on thin geometries, where side normals and sharp edges are difficult to recover from depth. Our model avoids this failure mode by complementing contact-based proposals with top-down grasps, which do not require estimated normals.~\footnote{We did not anticipate this result, as we initially included top-down grasp proposals only to increase data-collection diversity. We tried to improve the performance of EdgeGraspNet by changing viewpoint, filtering the pointcloud differently, and adjusting proposals after generation. However, we could not recover reliable grasps without either modifying the pointcloud normals directly or changing EdgeGraspNet's central assumption that grasps are parameterized by contact normals.} Nevertheless, our base model still fails in some cases. Some geometrically plausible grasps slip on low-friction surfaces or become unstable because the object has a non-uniform mass distribution. \green{In other cases, the geometric grasp sampler misses grasp modes when depth observations are incomplete.} 

\begin{figure}[t]
    \centering
    \setlength{\tabcolsep}{2pt}
    \begin{tabular}{ccc@{\hspace{0.5em}\vrule width 0.3pt\hspace{0.5em}}ccc@{\hspace{0.5em}\vrule width 0.3pt\hspace{0.5em}}ccc}
        \raisebox{-0.5\height}{\includegraphics[width=0.13\linewidth]{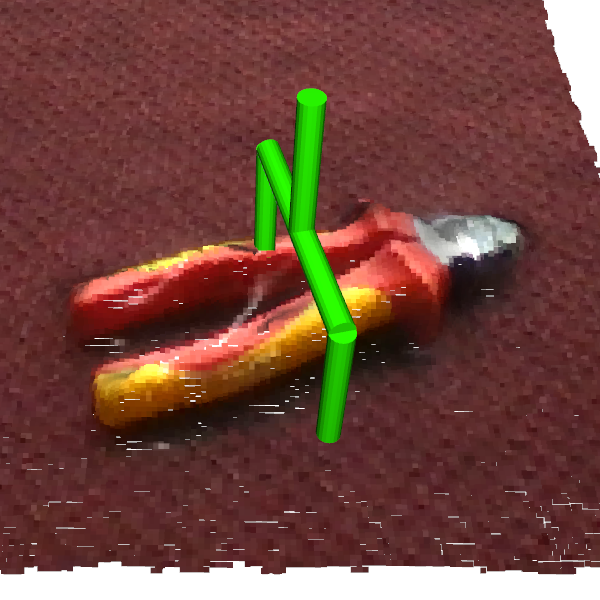}} &
        \(\boldsymbol{\scriptstyle\rightarrow}\) &
        \raisebox{-0.5\height}{\includegraphics[width=0.13\linewidth]{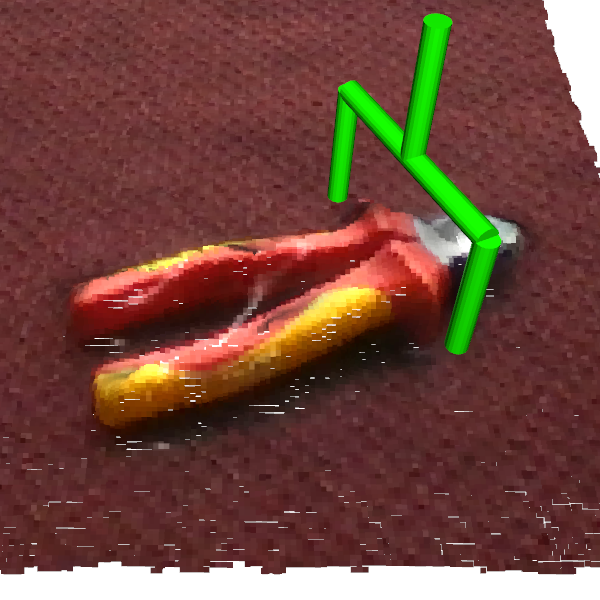}} &
        \raisebox{-0.5\height}{\includegraphics[width=0.13\linewidth]{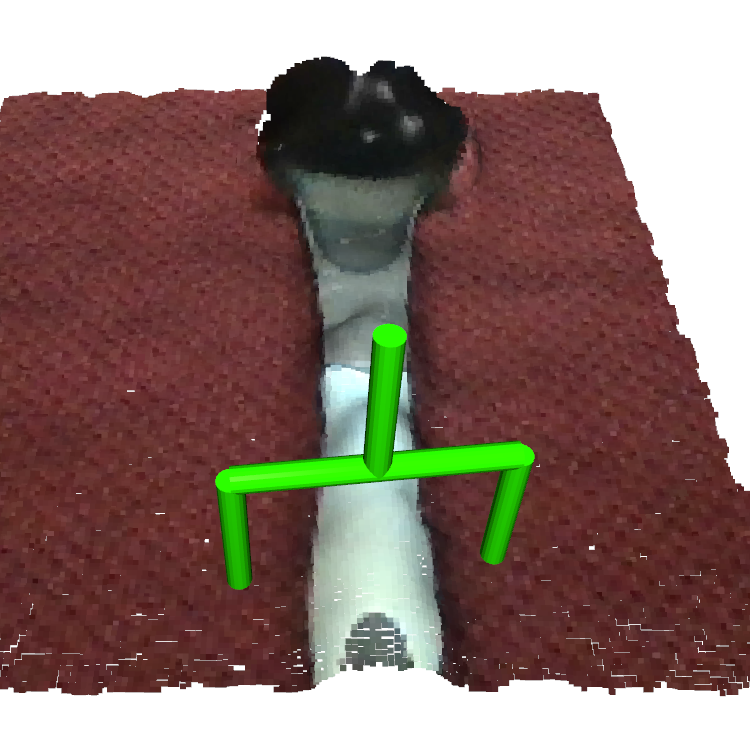}} &
        \(\boldsymbol{\scriptstyle\rightarrow}\) &
        \raisebox{-0.5\height}{\includegraphics[width=0.13\linewidth]{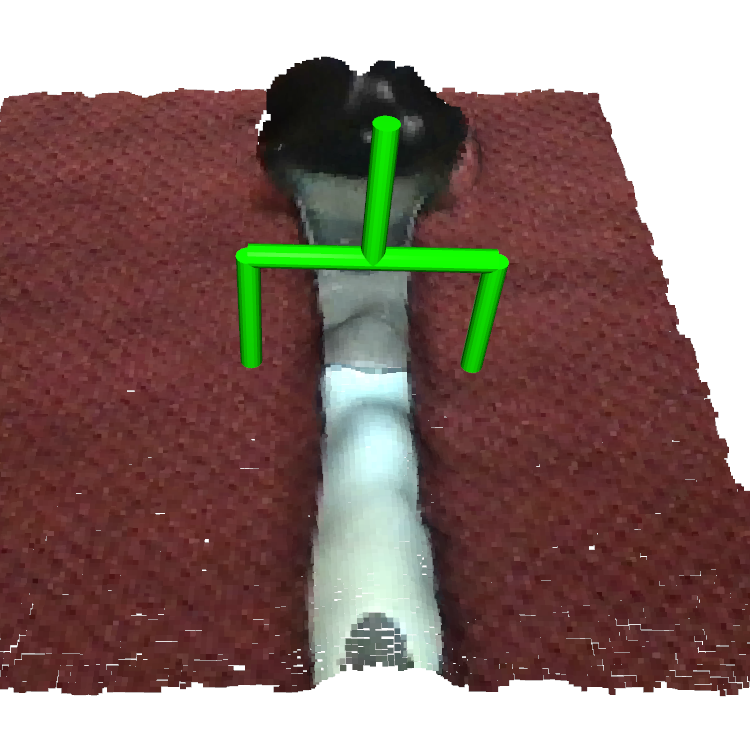}} &
        \raisebox{-0.5\height}{\includegraphics[width=0.13\linewidth]{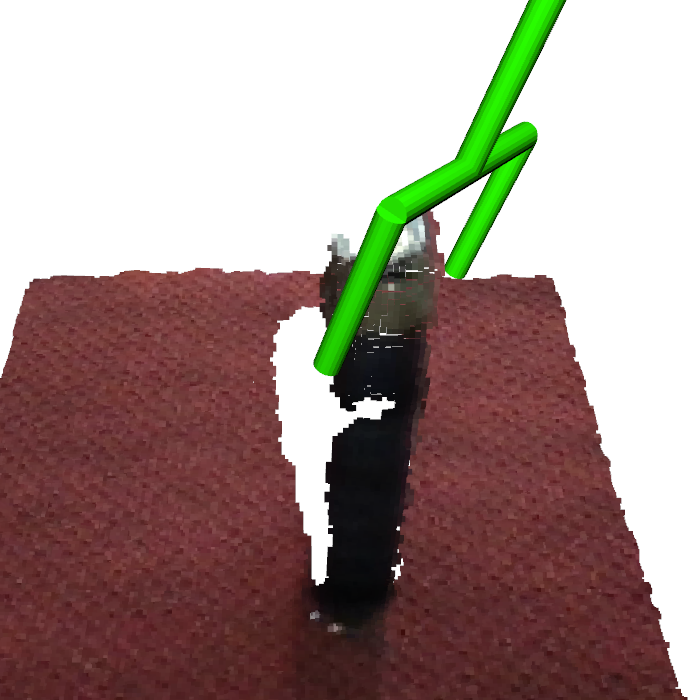}} &
        \(\boldsymbol{\scriptstyle\rightarrow}\) &
        \raisebox{-0.5\height}{\includegraphics[width=0.13\linewidth]{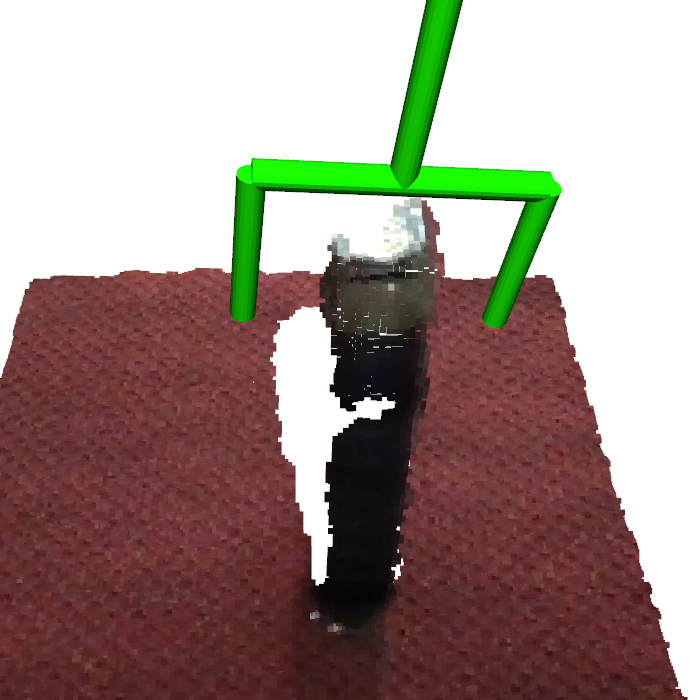}}
    \end{tabular}
\caption{
\textbf{Qualitative examples of real-world adaptation.}
    \green{On the pliers, repeated outcomes shift grasp preferences away from the low-friction handles and toward a grasp that locks against the metal head. On the ladle, they shift grasps closer to the center of mass. 
    On the shampoo bottle, the pointcloud misses much of the object body from the thin side, so the geometric sampler does not reliably generate grasps that wrap around the wider side. Demonstration recall adds this missing grasp mode to the proposal set.}
    }
    \label{fig:real_adaptation_examples}

\end{figure}

Online adaptation increases success rates across all categories, often substantially. \blue{After adaptation, our method achieves the highest success rate among all evaluated methods in every category and exceeds 90\% success in five of six categories.} \green{Qualitatively, the selected grasps change in ways consistent with the observed failures: repeated outcomes shift grasps away from low-friction areas and toward the center of mass, while demonstrations add grasp modes missing under partial observations.} 
We show some examples of this in~\figref{fig:real_adaptation_examples}. However, there remain cases where adaptation helps only partially. On the pliers category, success improves only from 62.5\% to 68.3\%. This is because one plier instance requires a grasp strategy disjoint from those that work on the others: grasps that work on the other pliers tend to fail on this one, and vice versa. Since these instances are difficult to distinguish from geometry alone, our system selects the strategy that works most reliably on average, which in turn causes repeated failures on this specific pliers instance. 

\paragraph{Continual Learning}
We construct 20 additional mixed scenes containing objects from all categories. We do not build this model by adapting to each category sequentially. Instead, we start from the category-specific models learned in the previous experiment, and merge their scoring and recall memories. We then evaluate this merged-memory model directly, without additional finetuning, consolidation, or corrective interaction. The final column of~\tableref{tab:real_category_adaptation} reports this mixed-scene evaluation: the merged-memory model reaches 89.6\% success and 100.0\% scene-clear rate, compared with 72.4\% and 85.0\% for the base model, \blue{69.2\% and 80.0\% for AnyGrasp, 50.0\% and 45.0\% for M2T2,} and 60.0\% and 30.0\% for EdgeGraspNet. This shows that independently collected scoring and recall memories can be combined without destructive interference in our real-world setting. It also points to a possible form of parallel adaptation: robots deployed in different environments, such as different households, could collect grasp outcomes and demonstrations independently, then pool the resulting memories.

\begin{table}[t]
\centering
\resizebox{\linewidth}{!}{
\begin{tabular}{lcccccc|lc}
\toprule
\multicolumn{7}{c|}{Single-category adaptation} & \multicolumn{2}{c}{Continual learning} \\
\cmidrule(lr){1-7}\cmidrule(lr){8-9}
Method & Control objects & Mugs and Bowls & Kitchen Tools & Pliers & Screwdriver & Toys & Method & Mixed scenes\\
\midrule
\blue{M2T2} & \blue{53.8\% (65.0\%)} & \blue{82.5\% (100\%)} & \blue{32.4\% (35.0\%)} & \blue{11.0\% (5.0\%)} & \blue{11.3\% (10.0\%)} & \blue{77.4\% (90.0\%)} & \blue{M2T2} & \blue{50.0\% (45.0\%)} \\
\blue{AnyGrasp} & \blue{78.8\% (95.0\%)} & \blue{83.3\% (95.0\%)} & \blue{71.9\% (90.0\%)} & \blue{28.6\% (30.0\%)} & \blue{54.9\% (90.0\%)} & \blue{85.0\% (100\%)} & \blue{AnyGrasp} & \blue{69.2\% (80.0\%)} \\

EdgeGraspNet & 80.2\% (90.0\%) & 88.5\% (\textbf{100\%}) & 37.8\% (20\%) & 0\% (0\%) & 0\% (0\%) & 72.5\% (85.0\%) & EdgeGraspNet & 60.0\% (30.0\%) \\
Ours (base) & 80.0\% (95.0\%) & 92.7\% (\textbf{100\%}) & 77.4\% (95.0\%) & 62.5\% (80.0\%) & 85.7\% (90.0\%) & 83.6\% (\textbf{100.0\%}) & Ours (base) & 72.4\% (85.0\%) \\
Ours (w adaptation) & \textbf{93.2\%} (\textbf{100\%}) & \textbf{100\%} (\textbf{100\%}) & \textbf{91.3\%} (\textbf{100.0\%}) & \textbf{68.3\%} (\textbf{90.0\%}) & \textbf{96.0\%} (\textbf{100.0\%}) & \textbf{94.4\%} (\textbf{100.0\%}) & Ours (merged) & \textbf{89.6\%} (\textbf{100.0\%}) \\
\bottomrule
\end{tabular}
}
\vspace{0.5em}
\caption{
\textbf{Real-world evaluation.}
Left block: success rate and scene-clear rate (in parentheses) on each real-object category, comparing \blue{M2T2, AnyGrasp, and} EdgeGraspNet against our base model and our model after independent category-level adaptation. Right block: performance on scenes containing objects from all categories, using a single model adapted to all categories. Online adaptation improves our model on every category and exceeds 90\% success rate on five of six categories. On mixed scenes, the continual-learning model reaches 89.6\% success and 100.0\% scene-clear rate. In total, our real-world experiments comprise more than 1500 grasp attempts.
}
\label{tab:real_category_adaptation}

\vspace{-2.0em}
\end{table}

\section{Limitations}
\label{sec:limitations}
 First, like other non-parametric methods, our approach must manage the memory and runtime cost of accumulated experience. In our system, the main cost is the runtime associated with demonstration recall. This was not a limiting factor in our experiments, even after adapting to hundreds of objects. For larger deployments (i.e. thousands of objects, hundreds of demonstrations), semantic filtering before recall could reduce this cost by recalling only relevant demonstrations. \blue{Appendix~\ref{appendix:runtime_memory} analyzes how the runtime and memory costs of our method scale with accumulated experience.}
Second, our pipeline relies on geometric observations alone. This can fail when geometry is not enough to distinguish similar objects that require different grasp strategies, as in the pliers failure case in our real-world experiments. Incorporating instance-level cues could help resolve these ambiguities.
Third, our adaptation strategy cannot recover geometry that is missing or distorted in the input observation. This limitation is not specific to our camera: depth sensors generally have failure modes, and reflective, transparent, or metallic objects remain challenging for many perception pipelines. In preliminary experiments, we tested learned depth enhancement models~\citep{liu2025manipulation}. Although they helped in some cases, they could also distort surfaces or absolute distances, introducing new errors for grasp planning.
Finally, \green{our experiments focus primarily on challenging objects and on object categories that are absent or underrepresented in the training data. We did not systematically evaluate other sources of deployment variation, such as sensor noise or changes in contact dynamics.}

\section{Conclusion}
\label{sec:conclusions}
In this work, we presented a continual-learning framework for single-view 6-DoF grasp synthesis in cluttered scenes. Our method augments a standard sample-and-score pipeline with memory: grasp outcomes update future scores, while user demonstrations add grasp modes that fixed heuristics may miss. Across simulation and real-world experiments, we show our system preserves strong offline performance while giving the robot a practical way to improve during deployment. In the real world, adaptation improves performance across every evaluated category and exceeds 90\% success rate in five of the six categories tested. Furthermore, we show that our system can keep accumulating experience over hundreds of new objects without forgetting previous adaptations. These results show that grasping systems need not stop learning once offline training is complete. By storing deployment outcomes and demonstrations in structured memories, a robot can turn failures, successes, and occasional user input into immediate changes in behavior, \green{enabling grasping policies that continue to improve during deployment}.

\acknowledgments{
This work was partially supported by the HILTI Group.

We acknowledge the use of artificial-intelligence-based tools as writing aids (editing and grammar enhancement) in the preparation of this manuscript. All intellectual content and conclusions remain the sole responsibility of the authors.
}

\bibliography{references}
\clearpage
\section*{Supplementary Material}

\def\graspingmainpaper{}
\ifdefined\graspingmainpaper
\appendix
\newcommand{\supplementarybibliography}{}
\newcommand{\supplementaryenddocument}{}
\else
\documentclass{article}

\usepackage{corl_2026}
\usepackage{graphicx}
\usepackage{subcaption}
\usepackage{siunitx}
\usepackage{amsmath}
\usepackage{amssymb}
\usepackage{algorithm}
\usepackage{algpseudocode}
\usepackage{booktabs}
\usepackage{tabularx}
\usepackage[normalem]{ulem}
\usepackage{placeins}
\usepackage{multirow}
\usepackage{pgfplots}
\pgfplotsset{compat=1.16}
\definecolor{revisionblue}{rgb}{0,0,1}
\definecolor{revisionred}{rgb}{1,0,0}
\definecolor{revisiongreen}{rgb}{0,1,0}
\definecolor{revisionwhite}{rgb}{1,1,1}
\definecolor{black}{rgb}{0,0,0}
\definecolor{mydarkblue}{rgb}{0,0.08,0.45}

\newcommand{\citationneeded}{\textcolor{revisionblue}{[citation needed]}}
\newcommand{\blue}[1]{\textcolor{revisionblue}{#1}}
\newcommand{\edit}[1]{#1}
\newcommand{\red}[1]{\textcolor{revisionred}{#1}}
\newcommand{\todo}[1]{\textcolor{revisionred}{#1}}
\newcommand{\green}[1]{\textcolor{revisiongreen}{#1}}
\newcommand{\white}[1]{\textcolor{revisionwhite}{#1}}
\newcommand{\figref}[1]{Figure \ref{#1}}
\newcommand{\algoref}[1]{Algorithm \ref{#1}}
\newcommand{\tableref}[1]{Table \ref{#1}}
\newcommand{\equationref}[1]{Equation~\ref{#1}}
\newcommand{\comment}[1]{\(\triangleright\) {#1}}
\newcommand{\delete}[1]{\sout{#1}}

\newtheorem{assumption}{Assumption}[section]
\captionsetup{font=small}

\begin{document}
\appendix
\newcommand{\supplementarybibliography}{\bibliography{references}}
\newcommand{\supplementaryenddocument}{\end{document}}
\fi

This supplementary material provides additional details and analyses that support the main paper. \blue{Appendix~\ref{appendix:evaluation_novelty} analyzes the relationship between the simulation training and evaluation distributions.} Appendix~\ref{appendix:grasp_sampling_heuristics} describes the grasp-frame convention and the geometric grasp proposal samplers. Appendix~\ref{appendix:heuristic_failure_cases} discusses failure cases of the contact-normal-based heuristic on thin objects. Appendix~\ref{appendix:embedding_details} gives implementation details for the memory-based scorer, including local patch encoding, encoder architecture, and scoring hyperparameters. \blue{Appendix~\ref{appendix:evidence_weights} details how evidence from different sources is weighted.} Appendix~\ref{appendix:demo_recall_details} describes how demonstrations are recalled as additional grasp proposals. Appendix~\ref{appendix:demonstration_interface} shows the simple Open3D interface used to provide demonstrations. Appendix~\ref{appendix:embedding_comparison} compares the SNN-trained embedding against a BCE-trained embedding. Appendix~\ref{appendix:forward_adaptation_sim} reports the full forward-adaptation results in simulation. Appendix~\ref{appendix:runtime_memory} reports runtime and memory costs before and after adaptation. \blue{Appendix~\ref{appendix:grasp_source_analysis} contains an analysis of the sources of executed grasps.} Finally, Appendix~\ref{appendix:real_demo_ablation} compares real-world adaptation with and without demonstrations.

\section{Training--Evaluation Distribution Analysis}
\label{appendix:evaluation_novelty}
\blue{We train in simulation using the object set shared by VGN, EdgeGraspNet, and ICGNet. For evaluation, we use 443 object instances from DexGraspNet that are not present in the training set. We attempted to choose categories that were underrepresented in training while still resembling objects a robot might encounter. Bowls, hammers, screwdrivers, and pliers occur in the training set only rarely, jointly accounting for less than 5\% of its objects, while airplanes and fasteners are absent. The remaining evaluation categories are present in the training set.}

\blue{To assess how far the evaluation distribution lies from the training distribution, we compare their grasp embeddings. We encode grasp proposals from the evaluation scenes with our learned encoder, compute the distance of each embedding to its fifth-nearest neighbor in the offline training memory, and report the median within each category. As an in-distribution reference, we apply the same procedure to held-out IID grasps on training objects.}

\begin{table}[h]
\centering
\resizebox{\linewidth}{!}{
\begin{tabular}{lccccccccccc}
\toprule
\blue{Split/category} & \blue{IID grasps} & \blue{Airplane} & \blue{Animal} & \blue{Bottle} & \blue{Bowl} & \blue{Drill} & \blue{Fasteners} & \blue{Hammer} & \blue{Mug} & \blue{Pliers} & \blue{Screwdriver} \\
\midrule
\blue{Median 5-NN distance} & \blue{0.36} & \blue{0.54} & \blue{0.56} & \blue{0.38} & \blue{0.48} & \blue{0.43} & \blue{0.27} & \blue{0.48} & \blue{0.59} & \blue{0.34} & \blue{0.41} \\
\bottomrule
\end{tabular}
}

\caption{\blue{Median fifth-nearest-neighbor distance from each category's grasp embeddings to the offline training memory. IID grasps consists of held-out grasps on training objects. Larger values indicate grasp-local geometry farther from the offline training memory.}}
\label{tab:evaluation_distribution_shift}
\end{table}

\blue{Most evaluation categories have a larger median distance than the IID reference, with airplanes, animals, bowls, hammers, and mugs showing the clearest shift. Bottles lie close to the IID reference, while fasteners and pliers have smaller distances. This indicates that several evaluation categories induce grasp geometries outside the training distribution, although the shift is not uniform across categories.}

\section{Grasp Parametrization and Grasp Proposal Sampling}
\label{appendix:grasp_sampling_heuristics}
We represent each grasp as a pose $(R,c) \in \mathrm{SE}(3)$, where $c \in \mathbb{R}^3$ is the gripper center and $R = [x\;y\;z] \in \mathrm{SO}(3)$ is the grasp frame. The axis $y$ is the finger closing direction, $z$ is the approach direction, and $x = y \times z$ completes the right-handed frame. We generate grasp proposals with two geometric heuristics.

\begin{figure}[h]
    \centering
    \includegraphics[width=0.5\linewidth]{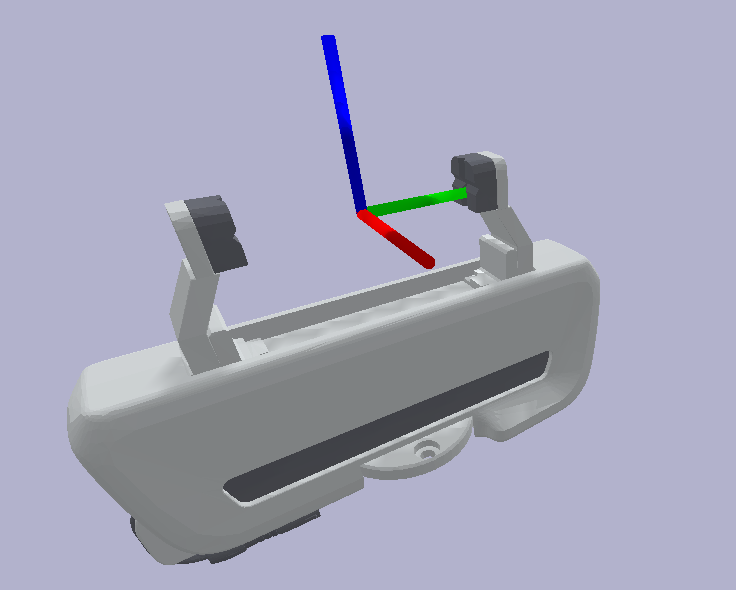}
    \caption{Grasp-frame convention used throughout the paper. The grasp frame is placed at the center of the two fingers. The y axis is defined by the finger closing direction. The z axis is the approach direction.}
    \label{fig:grasp_frame_convention}
\end{figure}

The first heuristic is inspired by EdgeGraspNet, ContactGraspNet, and ICGNet~\citep{edgegraspnet,contact_graspnet,icg_net}. We sample an approach point $p_a$ and a nearby contact point $p_c$, estimate the surface normal $n_c$ at $p_c$, set the closing direction to $y = n_c$, and define the approach direction as the normalized vector $z \propto n_c \times (n_c \times (p_a-p_c))$. The center is then placed as $c = p_a - \delta z$, where $\delta$ accounts for the gripper depth and the offset between $p_a$ and $p_c$ along $z$. This heuristic produces diverse side grasps, but depends on estimating $n_c$ reliably. \figref{fig:edge_failures} illustrates typical failure cases when thin side surfaces are not observed accurately.

We therefore also use a top-down sampler that does not rely on contact normals. We sample two nearby points $p_i$ and $p_j$ from the upper layer of the pointcloud, set the center to their midpoint $c = (p_i+p_j)/2$, define the closing direction as the normalized horizontal vector $y \propto (p_j-p_i)_{xy}$, and fix the approach direction to $z = (0,0,-1)$. We then set $x = y \times z$ and shift the center slightly along $z$ to vary insertion depth. This sampler is less expressive, but \green{does not require reliable side geometry or local surface normals}.

\section{Failure cases of heuristic grasp proposal}
\label{appendix:heuristic_failure_cases}
As detailed in the previous section, the contact-based proposal sampler parameterizes each candidate around an observed contact point and its estimated local surface normal. This is effective when the depth observation contains clean side surfaces, but it can fail on thin tabletop objects. These objects have only a small visible side area from the camera viewpoint, so the side geometry is often either smoothed into a sloped surface or missing from the pointcloud entirely.

\begin{figure}[h]
    \centering
    \begin{subfigure}[t]{0.31\linewidth}
        \centering
        \includegraphics[width=\linewidth]{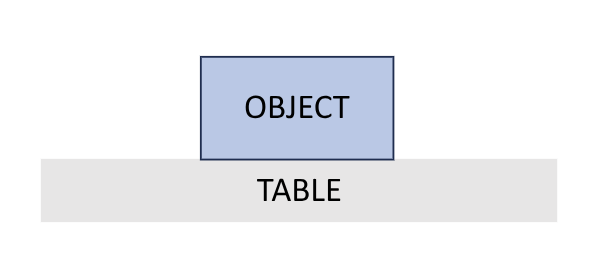}
        \caption{Actual scene geometry.}
    \end{subfigure}\hfill
    \begin{subfigure}[t]{0.32\linewidth}
        \centering
        \includegraphics[width=\linewidth]{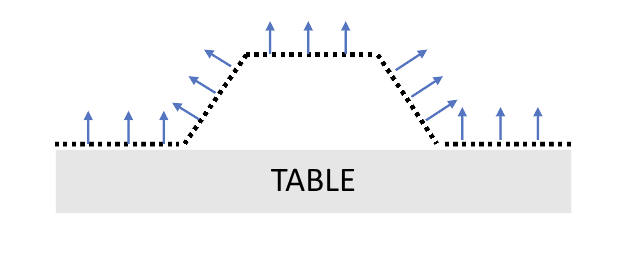}
        \caption{Sides smoothed into slopes.}
    \end{subfigure}\hfill
    \begin{subfigure}[t]{0.31\linewidth}
        \centering
        \includegraphics[width=\linewidth]{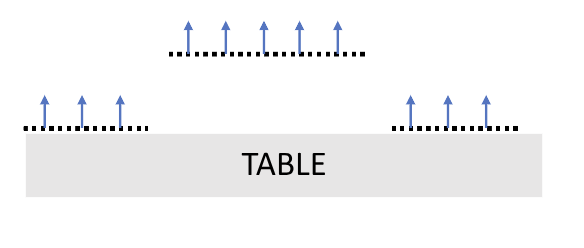}
        \caption{Sides missing from depth.}
    \end{subfigure}
    \caption{
    \textbf{Failure cases for contact-normal grasp proposals on thin objects.}
    The left panel shows the actual scene geometry: a thin object with vertical sides resting on the table. In real depth observations, these small side surfaces may be smoothed into slopes or absent from the pointcloud. The contact-based sampler then either follows an invalid sloped normal, causing finger-table collision, or has no side contacts from which to sample the grasp.
    }
    \label{fig:edge_failures}
\end{figure}

\figref{fig:edge_failures} summarizes the resulting failure modes. If the reconstructed side is smoothed into a slope, the estimated normal points diagonally. A grasp constructed normal to this slope can place the opposing finger below the object, causing a collision with the table and making the proposal invalid. If the side surface is missing, there are no side contact points from which the sampler can generate a grasp.

\green{These examples arise from naturally occurring artifacts in the real depth observations. We did not inject controlled levels of sensor noise, so they illustrate observed perception failure modes rather than a quantitative noise-robustness evaluation.}

\section{Implementation details: memory-based scorer}
\label{appendix:embedding_details}
This section gives the implementation details for the memory-based scorer. We first describe how local grasp patches are extracted and encoded, then give the encoder architecture and scoring hyperparameters. Finally, we describe how user demonstrations are weighted when they are added to the scoring memory.

\begin{figure}[h]
    \centering
    \includegraphics[width=\linewidth]{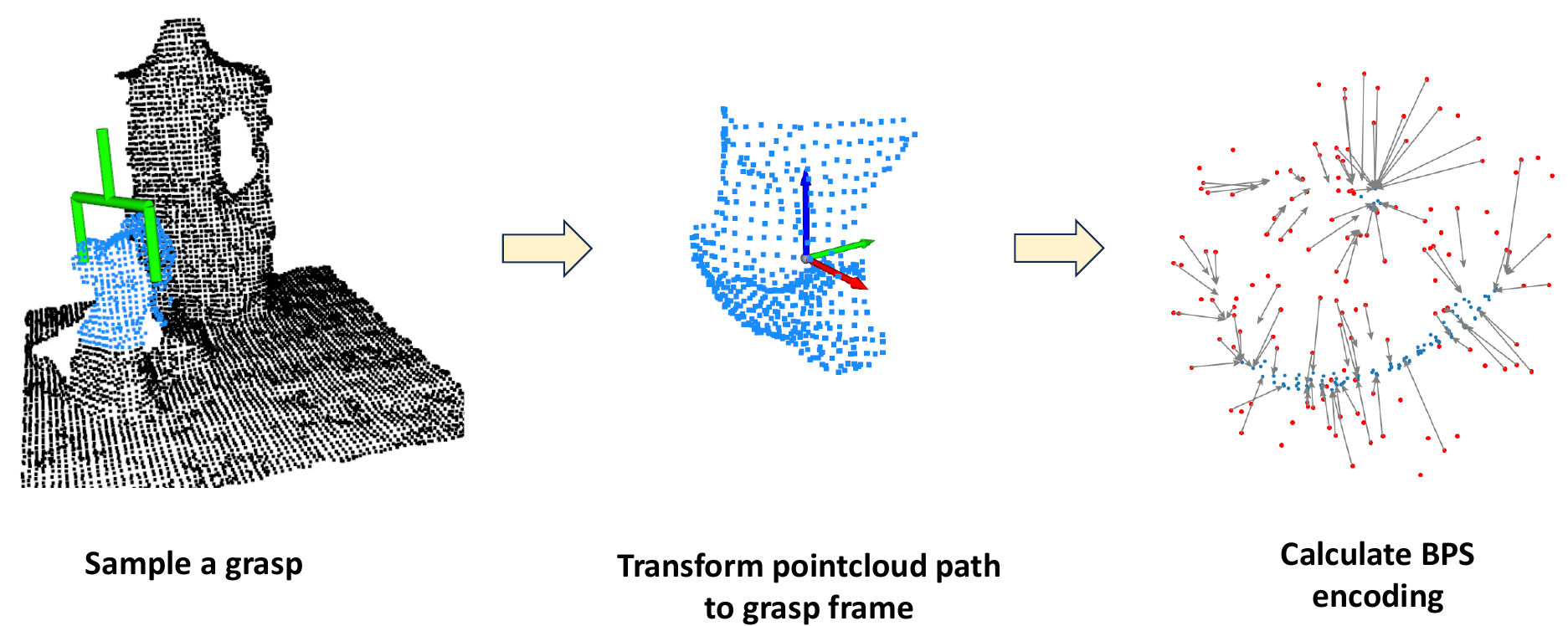}
    \caption{Local grasp patch extraction and Basis Point Set (BPS) encoding. Given a candidate grasp, we crop the local point cloud around the gripper, express it in the grasp frame, and encode it with fixed BPS basis points. For clarity, the figure visualizes a 2D slice of the BPS encoding. In the full representation, each basis point stores the displacement to its nearest observed scene point together with that point's surface normal.}
    \label{fig:embedding_patch_extraction}
\end{figure}

\paragraph{Patch extraction and basis point set encoding}
We crop a spherical patch with diameter 11 \si{cm} centered at the grasp pose and transform it in the grasp reference frame. Normals are estimated from the depth observation and transformed together with the points. We encode the aligned local patch with a spherical Basis Point Set representation. We use 512 fixed basis points sampled inside a sphere with the same dimension as the extracted patch. For each basis point, we find the nearest observed scene point and store the displacement vector to that point and the normal of that point. Finally, we serialize the representation to obtain a 6 $\times$ 512 descriptor, which is then used as input to the encoder MLP. \figref{fig:embedding_patch_extraction} shows the process.

\paragraph{Encoder and decoder.}
The serialized BPS features are passed to an MLP encoder with hidden dimensions 512, 256, and 128, which outputs a 32-dimensional embedding. The final embedding is then L2-normalized. A mirror MLP decoder reconstructs the BPS geometry from the embedding. At inference time, only the normalized encoder output is used by the continual-learning module.

\paragraph{Scorer hyperparameters.}
We retrieve $K=100$ nearest neighbors from memory for each query grasp. The SNN training temperature is $\tau_{\mathrm{train}}=0.2$. At inference time, the RBF kernel temperature $\tau_{\mathrm{eval}}$ is tuned to maximize average precision on validation data, which gives $\tau_{\mathrm{eval}}=0.3$. We use a Beta prior with $\alpha_0=0.1$ and $\beta_0=0.9$, and cap the total offline evidence at $W_{\max}^{\mathrm{off}}=10$.

\section{Tuning of Evidence Weights}
\label{appendix:evidence_weights}
\label{appendix:demo_weights}
\blue{We normalize the base evidence weights relative to offline data, setting $\rho_{\mathrm{off}}=1$. Online grasp outcomes use $\rho_{\mathrm{on}}=3$, so a nearby online observation contributes three times as much evidence as a nearby offline observation before distance-based attenuation. This ratio controls how quickly consistent deployment experience can revise the offline estimate.}

\blue{We cap the aggregate offline contribution to each query at $W_{\max}^{\mathrm{off}}=10$. Consequently, three mutually consistent online observations that are close to the query contribute up to nine pseudo-counts, which is comparable to the maximum evidence contributed by the entire local offline neighborhood. They can therefore substantially shift the posterior when they contradict the offline data, while a single online observation is not sufficient to dominate a strong offline estimate.}

\blue{Demonstrations are added as positive evidence with an adaptive base weight. To compute this weight, we first score the demonstrated grasp, obtaining posterior parameters $\alpha_d,\beta_d$, and the current best planner candidate, obtaining $\alpha_q^*,\beta_q^*$. We choose the demonstration weight so that adding it would raise the demonstrated grasp's odds above the current best candidate's odds by a margin $m_{\mathrm{demo}}$:}
\begin{equation}
\tilde{\rho}_d = \left(\frac{\alpha_q^*}{\beta_q^*} + m_{\mathrm{demo}}\right)\beta_d - \alpha_d,
\qquad
\rho_d = \max(m_{\mathrm{demo}}, \tilde{\rho}_d).
\label{eq:demo_weight}
\end{equation}
\blue{We use $m_{\mathrm{demo}}=1.0$ in all experiments. Intuitively, this means that if the same observation and candidate set were encountered again, the planner would select the demonstrated grasp, provided that it remains feasible.}

\section{Implementation details: demonstration recall}
\label{appendix:demo_recall_details}
As described in the main text, demonstration recall adds grasp proposals that may be missing from the geometric sampler. When we receive a demonstration, we store:
\begin{itemize}
    \item a local pointcloud patch around the demonstrated grasp, using a radius of $4\,\mathrm{cm}$. We compute FPFH descriptors and store them alongside the patch.
    \item the gripper pose (relative to the stored patch)
    \item the grasp embedding, obtained using the same encoder we use for the memory-based scorer
\end{itemize}

During inference, we use the FPFH features to match each stored patch against local neighborhoods in the current pointcloud. We sample up to 100 candidate neighborhood centers per demonstration, and rank them by the cosine similarity between their mean FPFH descriptor and the stored patch descriptor. Candidates with similarity below $0.7$ are discarded.

For each remaining candidate, we align the stored patch to the current neighborhood with point-to-plane ICP, initialized at the neighborhood centroid. The resulting transform maps the demonstrated grasp pose into the current scene, producing a recalled grasp proposal. We then encode each recalled grasp and keep only recalls whose feature is close to the original demonstration feature. Specifically, for each demonstration we keep recalls with
\[
    \lVert f_{\mathrm{rec}} - f_{\mathrm{demo}} \rVert_2 \le d_{\min} + 0.1,
\]
where $d_{\min}$ is the smallest feature distance among recalls from that demonstration in the current scene. 

This recall mechanism is deliberately simple, but we find that it works well in practice. We use it only to add proposals to the geometric sampler, not to choose the final grasp directly. Recalled grasps therefore still pass through the same collision filtering and memory-based scoring stages as all other proposals, which removes many poor matches. The use of a small local patch also helps recall generalize across similar objects, as reflected in the category-adaptation experiments. Nevertheless, recall is the part of our pipeline that would most benefit from future work. Local geometry alone is sometimes ambiguous, so more complex recall mechanisms could use semantics, instance recognition, or class-level templates in the future.

\section{Demonstration Interface}
\label{appendix:demonstration_interface}
\begin{figure}[h]
    \centering
    \begin{subfigure}[t]{0.48\linewidth}
        \centering
        \includegraphics[width=\linewidth]{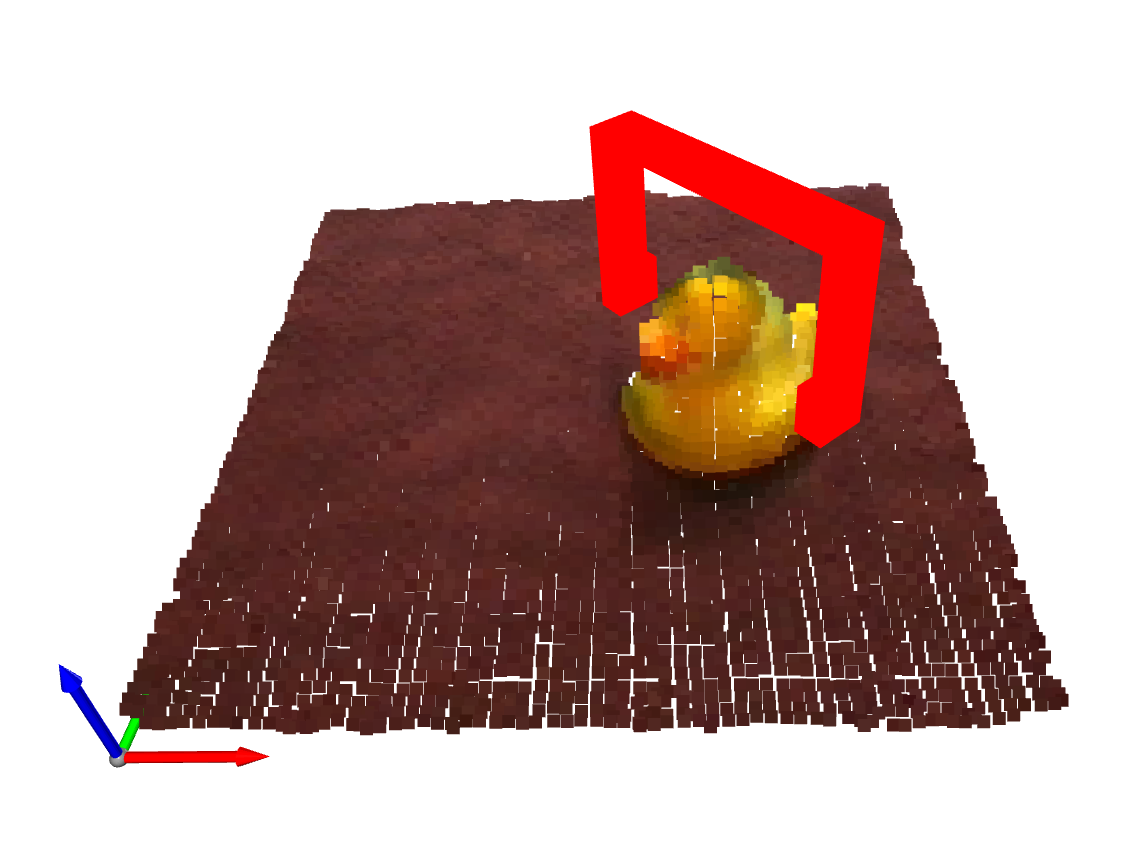}
        \caption{Rubber duck.}
        \label{fig:demonstration_interface_duck}
    \end{subfigure}
    \hfill
    \begin{subfigure}[t]{0.48\linewidth}
        \centering
        \includegraphics[width=\linewidth]{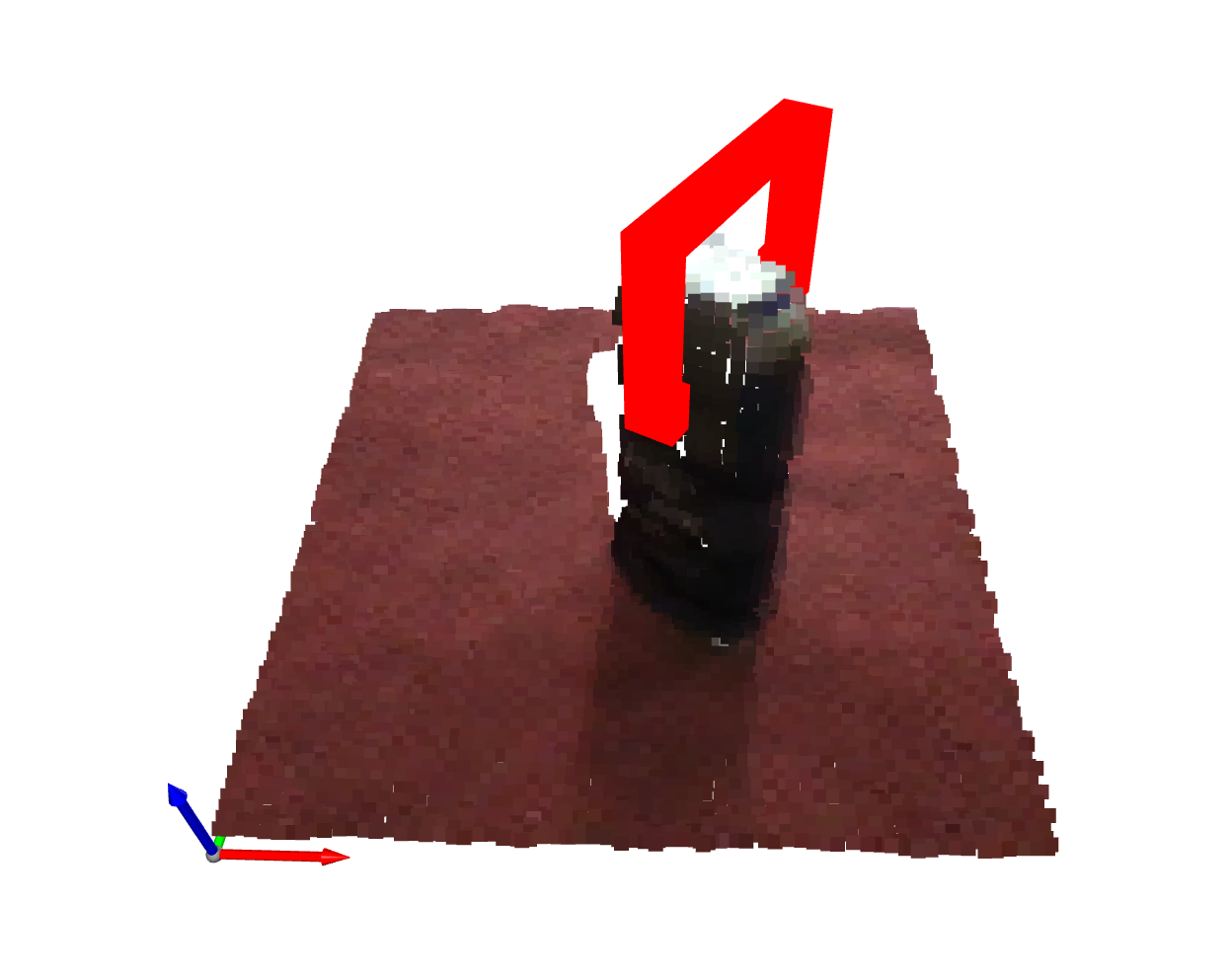}
        \caption{Shampoo bottle.}
        \label{fig:demonstration_interface_shampoo}
    \end{subfigure}
    \caption{
    \textbf{Open3D interface used to provide demonstrations.}
    The interface displays the observed point cloud and an editable gripper model. Users adjust the gripper pose with keyboard controls and confirm the demonstration by pressing enter.
    }
    \label{fig:demonstration_interface}
\end{figure}

We use a simple demonstration interface implemented in Open3D. The user is shown the robot's current observation together with a model of the parallel-jaw gripper, as shown in~\figref{fig:demonstration_interface}. The user moves the gripper pose with keyboard controls and presses enter to confirm the demonstration.

The resulting gripper pose is then added to the scoring memory and recall memory as described above.
Because the interface operates on the stored robot observation rather than requiring direct robot teleoperation, demonstrations can also be provided a posteriori. In principle, this means that a user could annotate failed or uncertain grasps after the robot has already collected the corresponding observations, without needing physical access to the robot at demonstration time.

\section{Comparison between SNN embedding and BCE embedding}
\label{appendix:embedding_comparison}
In our method, we train the low-dimensional grasp embedding with a soft nearest-neighbor loss and a reconstruction loss, following prior work on embeddings for non-parametric classifiers~\citep{hinton_snn_2007}. As an ablation, we also train an encoder with a standard binary-cross-entropy objective and use its normalized penultimate features as the embedding space. The scoring and online update rules are otherwise unchanged. This comparison tests whether the continual-learning behavior comes mainly from the non-parametric update rule, or whether the metric structure induced by the SNN objective is important.

\begin{table}[h]
\centering
\resizebox{\textwidth}{!}{
\begin{tabular}{lllllllllllll}
\toprule
Method & Train Obj. & Airplane & Animal & Bottle & Bowl & Drill & Fasteners & Hammer & Mug & Pliers & Screwdriver & \multicolumn{1}{c}{\shortstack{Average\\(excl. training)}} \\
\midrule

Ours (base) (BCE emb)  & - & 91.9\% & 94.3\% & 96.9\% & 93.8\% & 95.7\% & 96.8\% & 95.2\% & 90.6\% & 98.1\% & 93.8\% & 94.7\% \\

Ours (+ Airplane)
& - & 95.9\% & - & - & - & - & - & - & - & - & - & - \\

Ours (+ Animal)
& - & 94.7\% & 94.1\% & - & - & - & - & - & - & - & - & - \\

Ours (+ Bottle)
& - & 94.2\% & 95.5\% & 99.0\% & - & - & - & - & - & - & - & - \\

Ours (+ Bowl)
& - & 95.3\% & 95.7\% & 99.1\% & 98.6\% & - & - & - & - & - & - & - \\

Ours (+ Drill)
& - & 93.2\% & 95.0\% & 99.3\% & 98.3\% & 97.9\% & - & - & - & - & - & - \\

Ours (+ Fasteners)
& - & 96.0\% & 95.9\% & 98.7\% & 98.0\% & 97.3\% & 98.7\% & - & - & - & - & - \\

Ours (+ Hammer)
& - & 96.8\% & 94.8\% & 99.1\% & 98.7\% & 98.0\% & 98.8\% & 98.3\% & - & - & - & - \\

Ours (+ Mug)
& - & 94.4\% & 95.9\% & 98.8\% & 91.9\% & 96.5\% & 99.1\% & 95.8\% & 90.4\% & - & - & - \\

Ours (+ Pliers)
& - & 93.6\% & 94.1\% & 98.9\% & 95.4\% & 96.2\% & 99.0\% & 97.1\% & 93.4\% & 94.8\% & - & - \\

Ours (+ Screwdriver)
& - & 93.8\% & 94.0\% & 98.9\% & 94.0\% & 96.4\% & 99.0\% & 97.5\% & 92.8\% & 95.0\% & 97.1\% & 95.9\% \\

\bottomrule
\end{tabular}
}
\caption{
\textbf{Sequential continual learning with a BCE embedding.}
The model uses the same memory-based update rule as the main method, but replaces the SNN-trained embedding with embeddings from a binary classifier.
}
\label{tab:continual_ours_bce}
\end{table}

Table~\ref{tab:continual_ours_bce} shows that the BCE embedding gives similar base performance to the SNN embedding, suggesting that both encoders learn features that are useful for offline grasp scoring. The difference becomes clearer during sequential adaptation. The BCE embedding still supports some continual improvement, but it exhibits more interference between related categories. For example, after adapting to mugs, performance on bowls drops strongly. The last few categories also show weaker improvement after adaptation, suggesting that the embedding space becomes less useful as more online evidence is added.

To investigate this difference, we plot the cumulative explained variance of the PCA components of each embedding space in~\figref{fig:embedding_pca_comparison} below. 
\begin{figure}[h]
    \centering
    \begin{subfigure}[t]{0.4\linewidth}
        \centering
        \includegraphics[width=\linewidth]{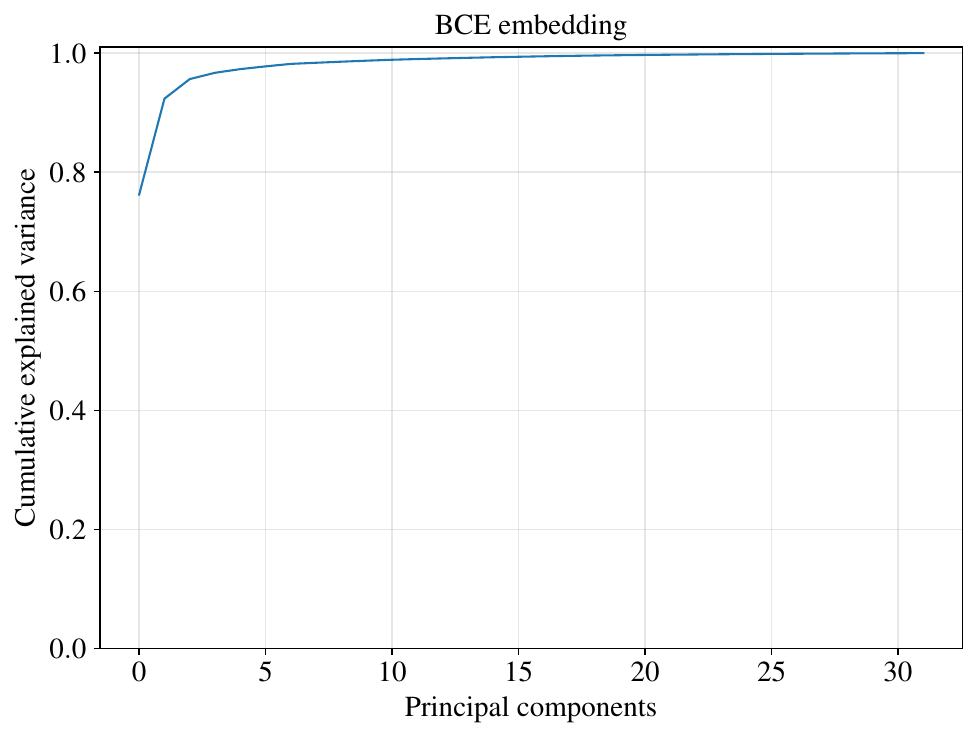}
        \caption{BCE embedding: cumulative explained variance.}
        \label{fig:bce_embedding_pca}
    \end{subfigure}
    \hfill
    \begin{subfigure}[t]{0.4\linewidth}
        \centering
        \includegraphics[width=\linewidth]{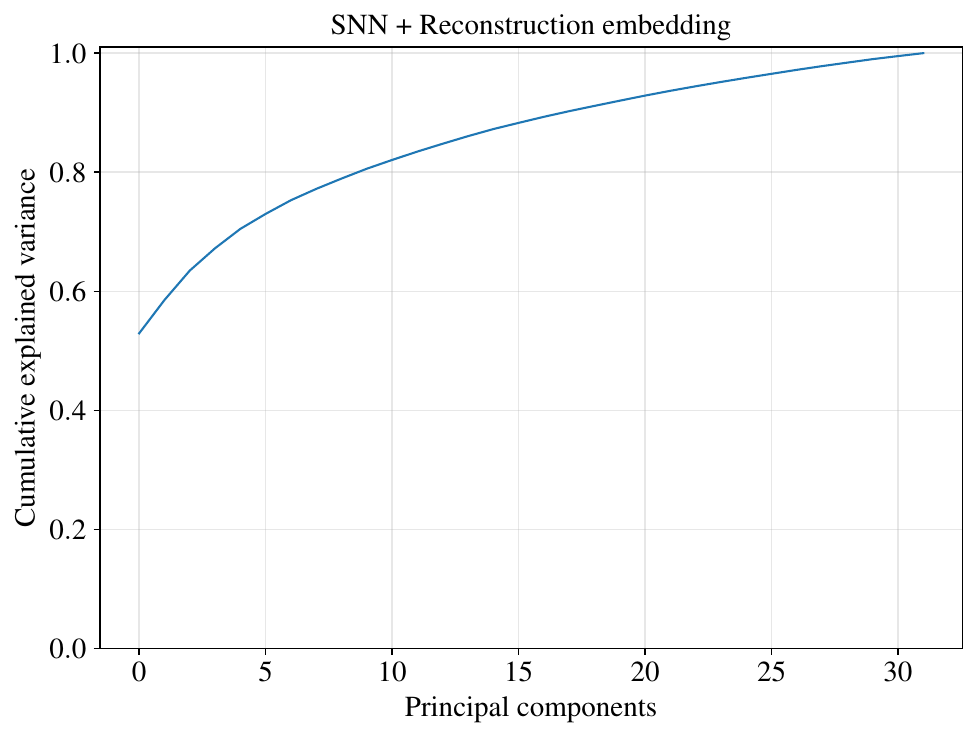}
        \caption{SNN+reconstruction embedding: cumulative explained variance.}
        \label{fig:snn_embedding_pca}
    \end{subfigure}
    \caption{
    \textbf{PCA cumulative-variance comparison of the learned embedding spaces.}
    Each plot shows the cumulative explained variance as a function of the number of principal components. The BCE embedding concentrates most variance in the first component, while the SNN+reconstruction embedding distributes variance across more dimensions, suggesting that it more effectively uses the embedding space to preserve geometry and separate multiple success and failure modes.
    }
    \label{fig:embedding_pca_comparison}
\end{figure}

The BCE objective concentrates variance along one dominant direction. This likely causes two effects. First, geometric details that are not directly useful for offline classification are discarded, which can later lead to interference between categories. Second, datapoints become concentrated in a small region of the embedding space, causing a saturation effect where new online evidence has limited influence. This could explain why the final adaptation categories show only limited improvement. By contrast, the SNN+reconstruction embedding spreads variance across more dimensions, preserving richer geometric structure and separating multiple success and failure modes. This can keep updates local to genuinely similar grasps and improve continual learning.

Nevertheless, we note that the pipeline still works reasonably reliably even with the BCE embedding. This is likely because we are not learning a sequence of strongly conflicting tasks, but refining graspability estimates in regions that were underrepresented during offline training. Therefore, even if the embedding is less well-structured, the online evidence can still improve the decision boundary without having to overwrite previously learned knowledge.

\section{Forward Adaptation in Simulation}
\label{appendix:forward_adaptation_sim}
Table~\ref{tab:forward_adaptation_sim} reports the full category-by-category performance matrix after adapting to each simulation category in the sequential continual-learning simulation experiment. Unlike the table in the main paper, this includes performance both on categories the model has seen already, and those it has not currently seen.

\begin{table}[h]
\centering
\resizebox{\textwidth}{!}{
\begin{tabular}{lllllllllllll}
\toprule
Method & Train Obj. & Airplane & Animal & Bottle & Bowl & Drill & Fasteners & Hammer & Mug & Pliers & Screwdriver & \multicolumn{1}{c}{\shortstack{Average\\(excl. training)}} \\
\midrule

Ours (base) & 98.1\% & 92.8\% & 93.8\% & 96.4\% & 88.6\% & 93.5\% & 97.4\% & 96.0\% & 92.4\% & 98.0\% & 97.2\% & 94.6\% \\

Ours (+ Airplane) & 97.8\% & 97.0\% & 96.2\% & 98.8\% & 96.3\% & 93.4\% & 96.5\% & 99.2\% & 92.0\% & 99.3\% & 99.2\% & 96.8\% \\

Ours (+ Animal) & 98.1\% & 96.2\% & 95.7\% & 97.8\% & 97.8\% & 93.1\% & 97.1\% & 98.5\% & 91.6\% & 98.5\% & 99.0\% & 96.5\% \\

Ours (+ Bottle) & 98.9\% & 96.2\% & 95.0\% & 99.5\% & 94.0\% & 94.0\% & 99.2\% & 99.2\% & 94.3\% & 98.3\% & 98.8\% & 96.8\% \\

Ours (+ Bowl) & 98.7\% & 96.8\% & 96.0\% & 99.5\% & 98.9\% & 96.1\% & 98.4\% & 99.1\% & 95.4\% & 98.7\% & 98.6\% & 97.8\% \\

Ours (+ Drill) & 98.7\% & 96.0\% & 95.0\% & 97.6\% & 98.3\% & 97.9\% & 98.9\% & 98.9\% & 94.3\% & 98.9\% & 98.9\% & 97.5\% \\

Ours (+ Fasteners) & 99.1\% & 96.5\% & 95.9\% & 98.7\% & 98.7\% & 98.6\% & 99.1\% & 98.7\% & 93.4\% & 98.6\% & 98.5\% & 97.7\% \\

Ours (+ Hammer) & 98.7\% & 95.2\% & 94.6\% & 98.5\% & 98.2\% & 98.1\% & 98.5\% & 98.0\% & 94.7\% & 99.1\% & 98.6\% & 97.4\% \\

Ours (+ Mug) & 98.2\% & 96.0\% & 93.3\% & 98.6\% & 96.9\% & 98.0\% & 98.7\% & 98.5\% & 99.2\% & 98.5\% & 98.6\% & 97.6\% \\

Ours (+ Pliers) & 98.7\% & 96.2\% & 94.8\% & 98.8\% & 97.5\% & 98.0\% & 98.8\% & 99.0\% & 99.0\% & 99.2\% & 98.1\% & 97.9\% \\

Ours (+ Screwdriver) & 98.1\% & 95.0\% & 94.5\% & 99.2\% & 98.0\% & 98.2\% & 98.4\% & 98.6\% & 98.3\% & 98.4\% & 99.1\% & 97.8\% \\

\bottomrule
\end{tabular}
}
\caption{
\textbf{Forward adaptation in simulation.}
\green{The robot adapts to the ten evaluation categories one after another.} After each stage, we evaluate on the current category and all categories (seen and unseen), using 1000 grasp attempts per category.
}
\label{tab:forward_adaptation_sim}
\end{table}

In general, we observe that adding new categories to the training data does not decrease performance on other unseen categories, \green{indicating that the improvement on some categories does not come at the cost of lower performance on the others}. Instead, there are some signs of forward adaptation: for example, adding the Animal category causes performance to increase for the Airplane and Bowl categories. Average performance on all categories climbs as more categories are added.

\section{Runtime and Memory Cost of Continual Learning}
\label{appendix:runtime_memory}
Because continual learning adds data throughout deployment, it is important to verify that adaptation does not make inference progressively slower or require impractical storage. We therefore measure the memory footprint and per-stage runtime before and after the sequential adaptation experiment. Timings are reported per planning cycle, from proposal generation through scoring. Runtime was measured on a machine equipped with an NVIDIA RTX 4090 GPU with 24 GB VRAM, a 12th Gen Intel Core i9-12900K CPU, and 32 GB RAM.

\begin{table}[h]
\centering
\resizebox{\linewidth}{!}{
\begin{tabular}{llllllll}
\toprule
Method & Mem. size (MB) & Sampling (ms) & Recall (ms) & Collision Check (ms) & Encoder (ms) & Scorer (ms) \\
\midrule
Ours (base) & 1.20 MB & 630 \si{ms} & - & 10 \si{ms} & 65 \si{ms} & 50 \si{ms} \\
Ours (end of adaptation) & 1.35 MB & 630 \si{ms} & 700 \si{ms} & 25 \si{ms} & 80 \si{ms} & 60 \si{ms} \\
\bottomrule
\end{tabular}
}
\caption{Runtime and memory before and after the sequential adaptation experiment. Timings are measured on a machine with 32 GB RAM, an NVIDIA GeForce RTX 4090, and an Intel Core i9-12900K.}
\label{tab:runtime_memory}
\end{table}

As shown in~\tableref{tab:runtime_memory}, the memory footprint remains small after adaptation. Adding new points to the memory-based scorer has little effect on either memory or runtime, since each new memory entry is only a low-dimensional embedding vector and a label. Most of the added cost comes from demonstration recall. We keep this bounded with a fixed 1000 ms recall budget, divided uniformly across demonstrations. For each demonstration, we stop once the time budget is exhausted or 30 grasp proposals are produced.

The fixed recall budget is a pragmatic choice. By default, we attempt to register each stored demonstration into the current scene, so without a budget the recall time would grow roughly linearly with the number of demonstrations. \blue{That is, uncapped recall has complexity $O(N_{\mathrm{demo}})$ in the number of stored demonstrations.} With the budget, runtime stays bounded, but each demonstration receives less registration time as the recall memory grows. In principle this could reduce recall quality after very long deployments. We do not observe such a drop in our experiments, even after adapting to hundreds of new objects. \blue{Adaptation required fewer than 50 demonstrations in total, so recall was not a bottleneck at the scale evaluated here.} It may become important at larger scales. A natural extension would be to filter demonstrations before registration, for example using semantic cues, so that recall spends its budget only on demonstrations likely to be relevant to the current scene. \blue{If object-class information were available, matching only demonstrations for the relevant class would reduce the per-scene work to $O(N_c)$, where $N_c$ is the number of demonstrations for that class.}

\section{Source of Executed Grasps}
\label{appendix:grasp_source_analysis}
\blue{For the adapted real-world model, we report the fraction of executed grasps originating from demonstration recall and from geometric sampling.}

\begin{table}[h]
\centering
\begin{tabular}{lcc}
\toprule
\blue{Object set} & \blue{Recall} & \blue{Geometric sampling} \\
\midrule
\blue{Control objects} & \blue{66\%} & \blue{34\%} \\
\blue{Mugs and bowls} & \blue{0\%} & \blue{100\%} \\
\blue{Kitchen tools} & \blue{59\%} & \blue{41\%} \\
\blue{Pliers} & \blue{10\%} & \blue{90\%} \\
\blue{Screwdrivers} & \blue{68\%} & \blue{32\%} \\
\blue{Toys} & \blue{39\%} & \blue{61\%} \\
\bottomrule
\end{tabular}
\caption{\blue{Source distribution of executed grasps after real-world adaptation. Values report the percentage of executed grasps originating from demonstration recall and geometric sampling.}}
\label{tab:real_grasp_sources}
\end{table}

\blue{The source distribution varies substantially across object sets. Mugs and bowls require no recalled grasps, whereas recall accounts for most executed grasps on the control objects, kitchen tools, and screwdrivers. Pliers and toys rely more heavily on geometric sampling. This variation supports using demonstration recall to supplement geometric sampling rather than replace it.}

\section{Real-World Adaptation With and Without Demonstrations}
\label{appendix:real_demo_ablation}
\begin{figure}[h]
    \centering
    \begin{subfigure}[t]{0.31\linewidth}
        \centering
        \includegraphics[height=2.8cm, width=\linewidth, keepaspectratio]{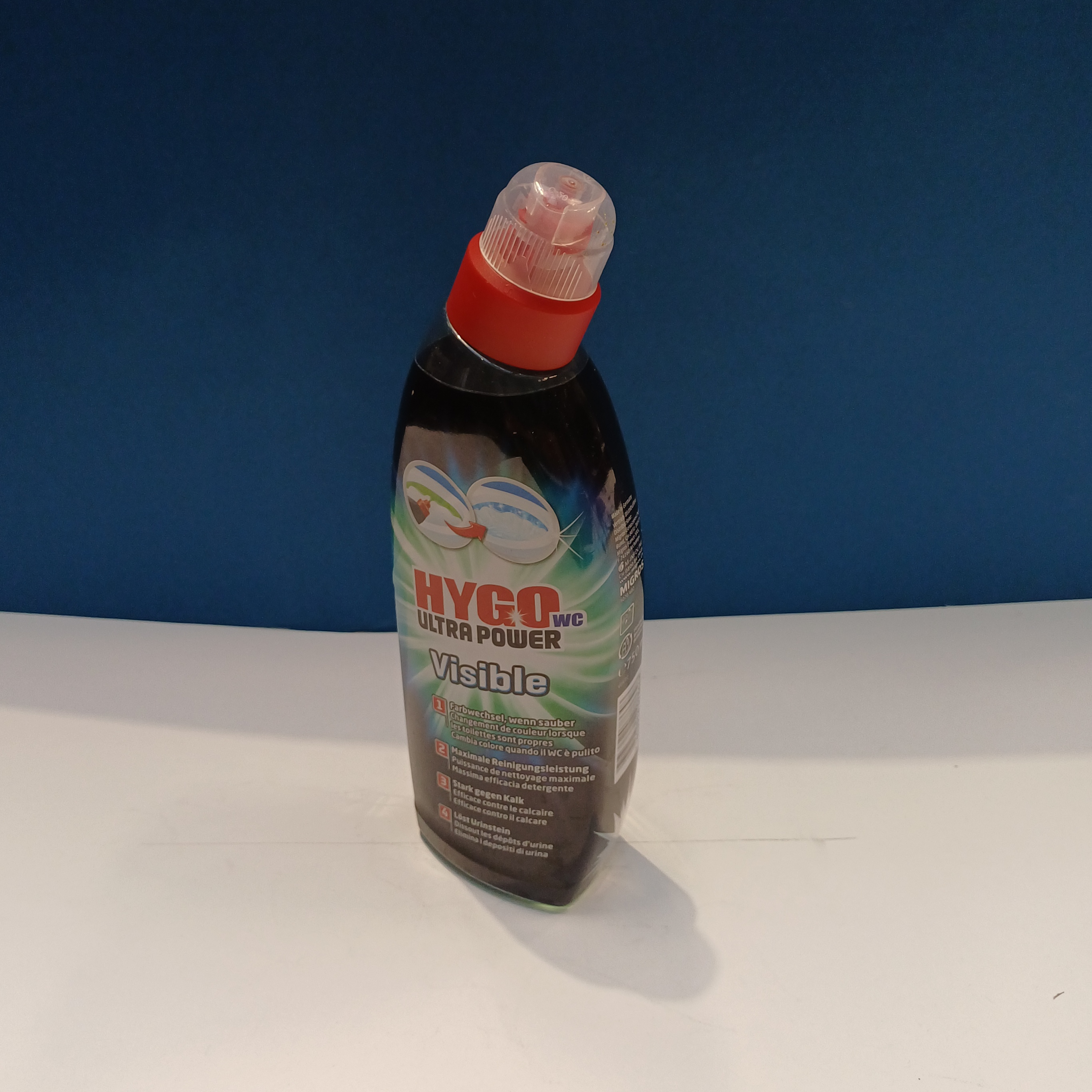}
        \caption{Toilet cleaner.}
        \label{fig:real_demo_ablation_object_toilet_cleaner}
    \end{subfigure}
    \hfill
    \begin{subfigure}[t]{0.31\linewidth}
        \centering
        \includegraphics[height=2.8cm, width=\linewidth, keepaspectratio]{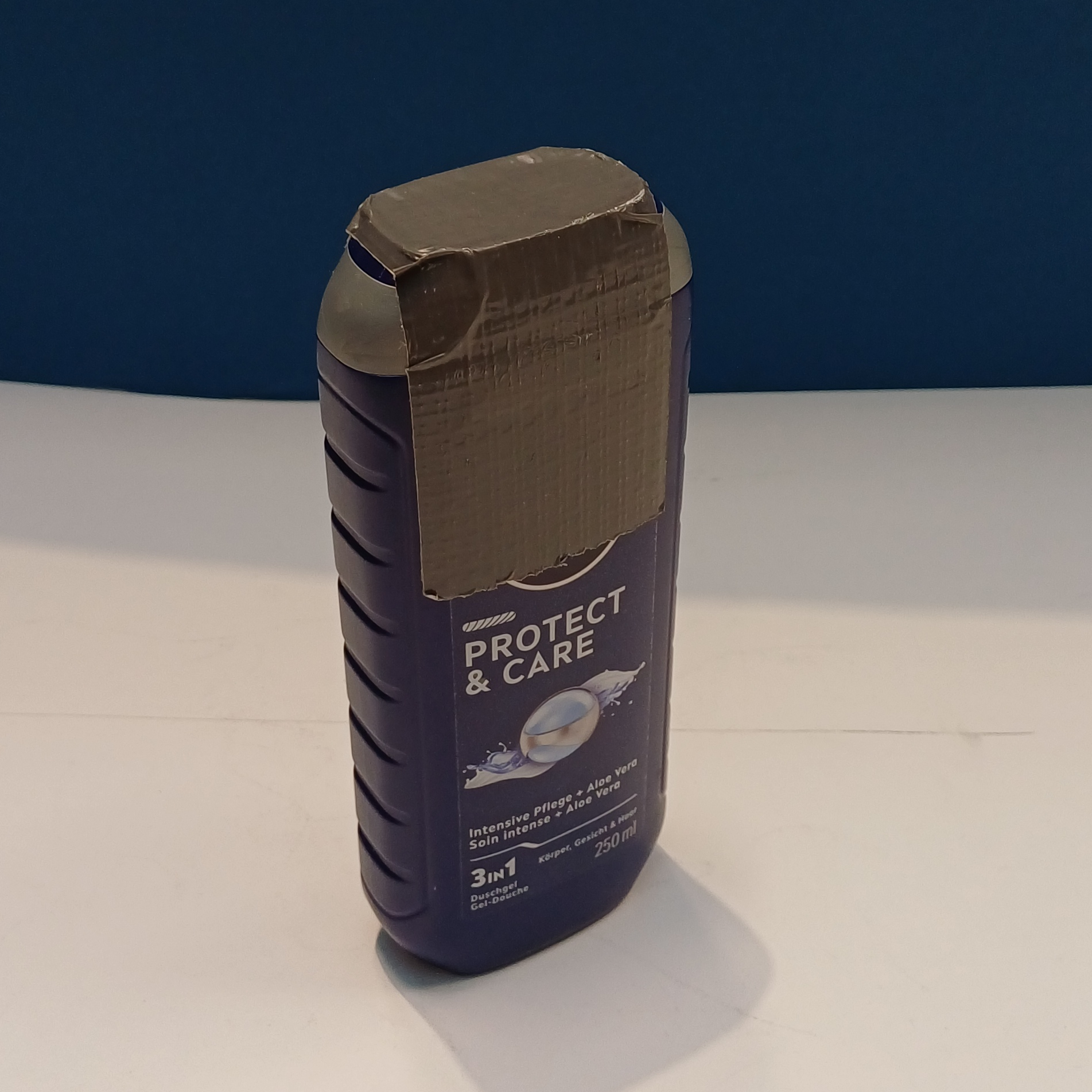}
        \caption{Shampoo bottle.}
        \label{fig:real_demo_ablation_object_shampoo}
    \end{subfigure}
    \hfill
    \begin{subfigure}[t]{0.31\linewidth}
        \centering
        \includegraphics[height=2.8cm, width=\linewidth, keepaspectratio]{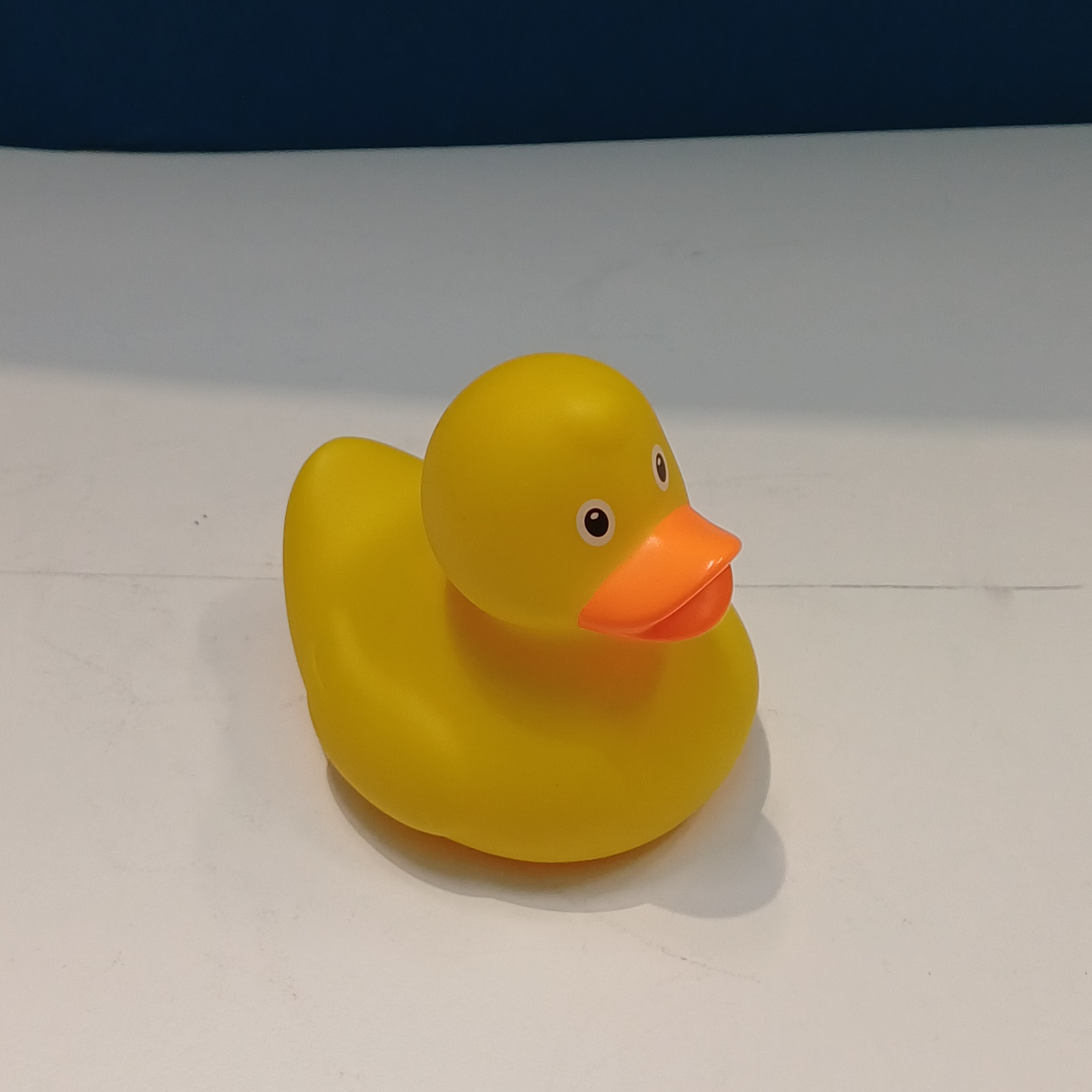}
        \caption{Rubber duck.}
        \label{fig:real_demo_ablation_object_rubber_duck}
    \end{subfigure}
    \caption{
    \textbf{Objects used for the real-world demonstration comparison.}
    The objects were selected from real-world cases that were difficult for the robot before adaptation.
    }
    \label{fig:real_demo_ablation_objects}
\end{figure}

While in simulation we carry out ablations that isolate the contributions of grasp outcomes and demonstrations, in our real-world experiments we use our full method: grasp outcomes are added to the scoring memory, and user demonstrations are added to the scoring and recall memories. Here we add a smaller real-world comparison to illustrate how demonstrations affect adaptation on individual difficult objects. We select three objects that were difficult to grasp for our base policy in the real-world trials: a toilet-cleaner bottle, a shampoo bottle, and a rubber duck (\figref{fig:real_demo_ablation_objects}). For each object, the robot starts from the base model and receives 20 adaptation attempts. We compare two variants: outcome-only adaptation, which updates the scoring memory from the binary grasp outcomes, and full adaptation, which is additionally aided by user demonstrations. For the full method, we initialize the demonstration memory with two demonstrations and then provide a new demonstration after each failed grasp.

\figref{fig:real_demo_ablation_trials} shows the resulting traces. The outcome-only variant needs more failed attempts before stabilizing on the toilet cleaner and shampoo bottle. Demonstrations reduce the number of failures from 7 to 3 on the toilet cleaner and from 5 to 2 on the shampoo bottle. On the rubber duck, demonstrations reduce failures from 6 to 4, although they do not remove all late failures. In both variants, failures generally become sparser over the 20 attempts, reflecting adaptation from accumulated experience. These traces are not intended as a comprehensive real-world benchmark, but they are consistent with the mechanism observed elsewhere in the paper: outcome feedback can improve scoring over repeated trials, while demonstrations can accelerate adaptation when a useful grasp mode is missing or underweighted. In this sense, demonstrations can trade a small amount of user input for fewer robot failures, which may be important in settings where failed grasps are costly.

\newcommand{\RealDemoTracePlot}[3]{
\begin{tikzpicture}
\begin{axis}[
    ybar,
    width=0.95\linewidth,
    height=3.0cm,
    ymin=0,
    ymax=1.15,
    xmin=0.5,
    xmax=20.5,
    bar width=2pt,
    enlarge x limits=false,
    ylabel={Failure},
    xlabel={Trial},
    xtick={1,2,3,4,5,6,7,8,9,10,11,12,13,14,15,16,17,18,19,20},
    xticklabel style={font=\scriptsize},
    ytick={0,1},
    yticklabels={0,1},
    tick label style={font=\scriptsize},
    label style={font=\scriptsize},
    title={#1},
    title style={font=\small},
    grid=major,
    major grid style={draw=black!10},
]
\addplot+[draw=none, fill=black!65, bar shift=-2pt] coordinates {#2};
\addplot+[draw=none, fill=blue!65, bar shift=2pt] coordinates {#3};
\end{axis}
\end{tikzpicture}
}

\newcommand{\RealDemoLegend}{
\small
\raisebox{0.2ex}{\tikz\draw[fill=black!65, draw=none] (0,0) rectangle (0.45em,0.9em);} Ours (no demonstrations)
\hspace{1.5em}
\raisebox{0.2ex}{\tikz\draw[fill=blue!65, draw=none] (0,0) rectangle (0.45em,0.9em);} Ours (full)
}

\begin{figure}[h]
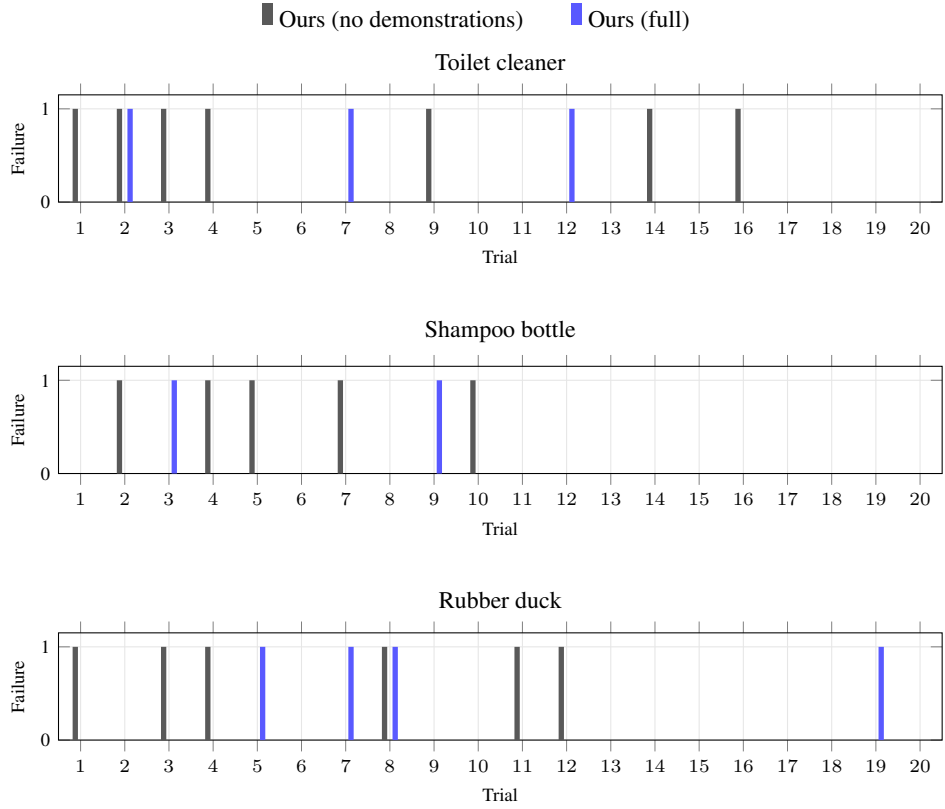

    \centering
    \RealDemoLegend
    \vspace{0.5em}

    \begin{subfigure}[t]{\linewidth}
        \centering
        \RealDemoTracePlot{Toilet cleaner}
        {(1,1) (2,1) (3,1) (4,1) (5,0) (6,0) (7,0) (8,0) (9,1) (10,0) (11,0) (12,0) (13,0) (14,1) (15,0) (16,1) (17,0) (18,0) (19,0) (20,0)}
        {(1,0) (2,1) (3,0) (4,0) (5,0) (6,0) (7,1) (8,0) (9,0) (10,0) (11,0) (12,1) (13,0) (14,0) (15,0) (16,0) (17,0) (18,0) (19,0) (20,0)}
        \label{fig:real_demo_ablation_trials_toilet_cleaner}
    \end{subfigure}
    \vspace{0.5em}

    \begin{subfigure}[t]{\linewidth}
        \centering
        \RealDemoTracePlot{Shampoo bottle}
        {(1,0) (2,1) (3,0) (4,1) (5,1) (6,0) (7,1) (8,0) (9,0) (10,1) (11,0) (12,0) (13,0) (14,0) (15,0) (16,0) (17,0) (18,0) (19,0) (20,0)}
        {(1,0) (2,0) (3,1) (4,0) (5,0) (6,0) (7,0) (8,0) (9,1) (10,0) (11,0) (12,0) (13,0) (14,0) (15,0) (16,0) (17,0) (18,0) (19,0) (20,0)}
        \label{fig:real_demo_ablation_trials_shampoo}
    \end{subfigure}
    \vspace{0.5em}

    \begin{subfigure}[t]{\linewidth}
        \centering
        \RealDemoTracePlot{Rubber duck}
        {(1,1) (2,0) (3,1) (4,1) (5,0) (6,0) (7,0) (8,1) (9,0) (10,0) (11,1) (12,1) (13,0) (14,0) (15,0) (16,0) (17,0) (18,0) (19,0) (20,0)}
        {(1,0) (2,0) (3,0) (4,0) (5,1) (6,0) (7,1) (8,1) (9,0) (10,0) (11,0) (12,0) (13,0) (14,0) (15,0) (16,0) (17,0) (18,0) (19,1) (20,0)}
        \label{fig:real_demo_ablation_trials_rubber_duck}
    \end{subfigure}
    \caption{
    \textbf{Real-world adaptation traces with and without demonstrations.}
    Each panel shows 20 adaptation attempts on one object, starting from the same base model. Bars indicate failed grasp attempts; missing bars indicate successful attempts. The black trace adapts only from binary grasp outcomes, while the blue trace uses the full method with demonstrations. For the full method, the demonstration memory is initialized with two demonstrations, and a new demonstration is added after each failed grasp.
    }
    \label{fig:real_demo_ablation_trials}
\end{figure}

\FloatBarrier

\supplementarybibliography
\supplementaryenddocument

\end{document}